\documentclass[journal]{IEEEtran}

\usepackage{amsmath,amssymb}
\usepackage{graphicx}
\usepackage[justification=centering]{caption}
\usepackage{color}
\usepackage[american]{babel}
\usepackage[caption=false, font=footnotesize]{subfig}
\usepackage{makecell}
\usepackage{setspace}
\usepackage{tabularx}
\usepackage{hyperref}

\DeclareMathOperator*{\argmin}{arg\,min}

\begin{document}

\title{Deep Learning for Single Image Super-Resolution: A Brief Review}

\author{Wenming~Yang, Xuechen~Zhang, Yapeng~Tian, Wei~Wang, Jing-Hao~Xue, Qingmin~Liao
\thanks{This work was partly supported by the National Natural Science Foundation of China (Nos.61471216 and 61771276), the National Key Research and Development Program of China (No.2016YFB0101001) and the Special Foundation for the Development of Strategic Emerging Industries of Shenzhen (Nos. JCYJ20170307153940960 and JCYJ20170817161845824). (Corresponding author: Wenming Yang)}
\thanks {W.~Yang, X.~Zhang, W.~Wang and Q.~Liao are with the Department of Electronic Engineering, Graduate School at Shenzhen, Tsinghua University, China (E-mail: \{yang.wenming@sz, xc-zhang16@mails, wangwei17@mails, liaoqm@\}.tsinghua.edu.cn.}
\thanks {Y.~Tian is with the University of Rochester, USA (E-mail: ytian21@ur.rochester.edu).}
\thanks {J.-H.~Xue is with the Department of Statistical Science, University College London, UK (E-mail: jinghao.xue@ucl.ac.uk).}
}


\maketitle

\begin{abstract} 
Single image super-resolution (SISR) is a notoriously challenging ill-posed problem that aims to obtain a high-resolution (HR) output from one of its low-resolution (LR) versions. Recently, powerful deep learning algorithms have been applied to SISR and have achieved state-of-the-art performance. In this survey, we review representative deep learning-based SISR methods and group them into two categories according to their contributions to two essential aspects of SISR: the exploration of efficient neural network architectures for SISR and the development of effective optimization objectives for deep SISR learning. For each category, a baseline is first established, and several critical limitations of the baseline are summarized. Then, representative works on overcoming these limitations are presented based on their original content, as well as our critical exposition and analyses, and relevant comparisons are conducted from a variety of perspectives. Finally, we conclude this review with some current challenges and future trends in SISR that leverage deep learning algorithms. 
\end{abstract}

\begin{IEEEkeywords}
Single image super-resolution, deep learning, neural networks, objective function
\end{IEEEkeywords}

\IEEEpeerreviewmaketitle


\section{Introduction}\label{s:intro}
\IEEEPARstart{D}{eep} learning (DL)~\cite{lecun2015deep} is a branch of machine learning algorithms that aims at learning the hierarchical representations of data. Deep learning has shown prominent superiority over other machine learning algorithms in many artificial intelligence domains, such as computer vision~\cite{krizhevsky2012imagenet}, speech recognition~\cite{hinton2012deep}, and natural language processing~\cite{collobert2008unified}. Generally, the strong capacity of DL to address substantial unstructured data is attributable to two main contributors: the development of efficient computing hardware and the advancement of sophisticated algorithms.
 
Single image super-resolution (SISR) is a notoriously challenging ill-posed problem because a specific low-resolution (LR) input can correspond to a crop of possible high-resolution (HR) images, and the HR space (in most instances, it refers to the natural image space) that we intend to map the LR input to is usually intractable~\cite{yang2014single}. Previous methods for SISR mainly have two drawbacks: one is the unclear definition of the mapping that we aim to develop between the LR space and the HR space, and the other is the inefficiency of establishing a complex high-dimensional mapping given massive raw data. Benefiting from the strong capacity of extracting effective high-level abstractions that bridge the LR and HR space, recent DL-based SISR methods have achieved significant improvements, both quantitatively and qualitatively.

In this survey, we attempt to give an overall review of recent DL-based SISR algorithms. We mainly focus on two areas: efficient neural network architectures designed for SISR and effective optimization objectives for DL-based SISR learning. The reason for this taxonomy is that when we apply DL algorithms to tackle a specified task, it is best for us to consider both the universal DL strategies and the specific domain knowledge. From the perspective of DL, although many other techniques such as data preprocessing~\cite{timofte2016seven} and model training techniques are also quite important~\cite{kingma2014adam,he2015delving}, the combination of DL and domain knowledge in SISR is usually the key to success and is often reflected in the innovations of neural network architectures and optimization objectives for SISR. In each of these two focused areas, based on the benchmark, several representative works are discussed mainly from the perspective of their contributions and experimental results as well as our comments and views. 

The rest of the paper is arranged as follows. In Section~\ref{s:back}, we present relevant background concepts of SISR and DL. In Section~\ref{s:arch}, we survey the literature on exploring efficient neural network architectures for various SISR tasks. In Section~\ref{s:obj}, we survey the studies on proposing effective objective functions for different purposes. In Section~\ref{s:trend}, we summarize some trends and challenges for DL-based SISR. We conclude this survey in Section~\ref{s:concl}.

\section{Background}\label{s:back}

\subsection{Single Image Super-Resolution}
\begin{figure}
    \centering
    \subfloat{
        \includegraphics[scale=0.3]{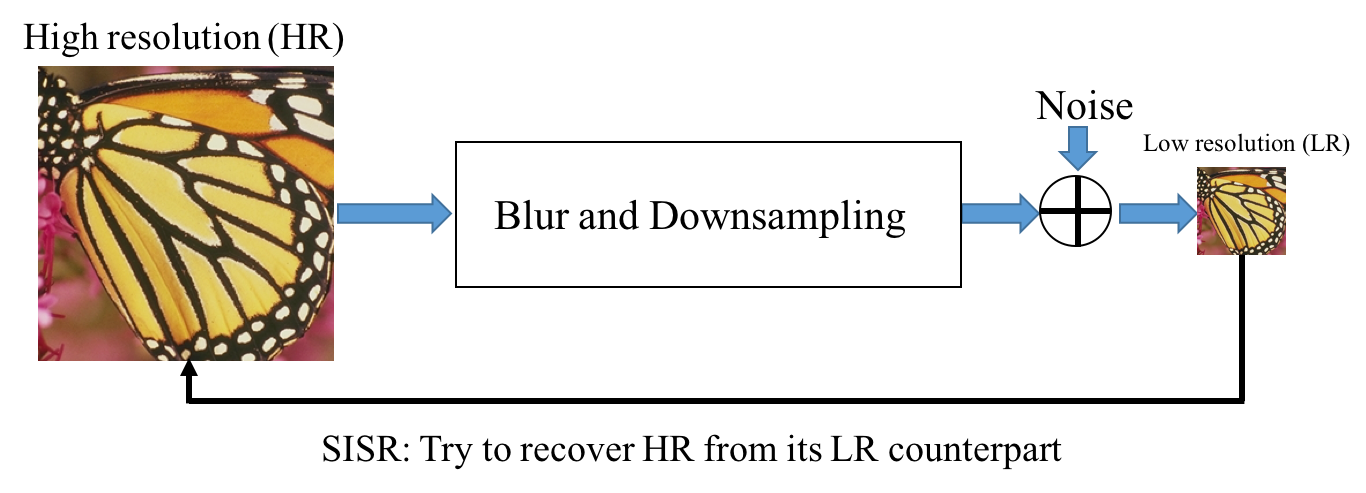}}
    \caption{Sketch of the overall framework of SISR.}
    \label{fig:introSISR}
\end{figure}

Super-resolution (SR)~\cite{park2003super} refers to the task of restoring high-resolution images from one or more low-resolution observations of the same scene. According to the number of input LR images, the SR can be classified into single image super-resolution (SISR) and multi-image super-resolution (MISR). 
Compared with MISR, SISR is much more popular because of its high efficiency. Since an HR image with high perceptual quality has more valuable details, it is widely used in many areas, such as medical imaging, satellite imaging and security imaging. In the typical SISR framework, as depicted in Fig.~\ref{fig:introSISR}, the LR image y is modeled as follows:
\begin{equation}
    \begin{split}
        y = (x \otimes k){\downarrow}_{s} + n,
    \end{split}
    \label{defSISR}
\end{equation}
where $x \otimes k$ is the convolution between the blurry kernel $k$ and the unknown HR image $x$, ${\downarrow}_{s}$ is the downsampling operator with scale factor $s$, and $n$ is the independent noise term. Solving (\ref{defSISR}) is an extremely ill-posed problem because one LR input may correspond to many possible HR solutions. To date, mainstream algorithms of SISR are mainly divided into three categories: interpolation-based methods, reconstruction-based methods and learning-based methods. 

Interpolation-based SISR methods, such as bicubic interpolation~\cite{keys1981cubic} and Lanczos resampling~\cite{duchon1979lanczos}, are very speedy and straightforward but suffer from accuracy shortcomings. Reconstruction-based SR methods~\cite{dai2009softcuts,sun2008image,yan2015single,marquina2008image} often adopt sophisticated prior knowledge to restrict the possible solution space with an advantage of generating flexible and sharp details. However, the performance of many reconstruction-based methods degrades rapidly when the scale factor increases, and these methods are usually time-consuming. 

Learning-based SISR methods, also known as example-based methods, are brought into focus because of their fast computation and outstanding performance. These methods usually utilize machine learning algorithms to analyze statistical relationships between the LR and its corresponding HR counterpart from substantial training examples. The Markov random field (MRF)~\cite{freeman2002example} approach was first adopted by 
Freeman \emph{et~al.} to exploit the abundant real-world images to synthesize visually pleasing image textures. Neighbor embedding methods~\cite{chang2004super} proposed by Chang \emph{et~al.} took advantage of similar local geometry between LR and HR to restore HR image patches. Inspired by the sparse signal recovery theory~\cite{aharon2006k}, researchers applied sparse coding methods~\cite{yang2010image,zeyde2010single,timofte2013anchored,timofte2014a+,cao2016image,liu2017retrieval} to SISR problems. Lately, random forest~\cite{schulter2015fast} has also been used to achieve improvement in the reconstruction performance. Meanwhile, many works combined the merits of reconstruction-based methods with the learning-based approaches to further reduce artifacts introduced by external training examples~\cite{zhang2017coarse,yu2014unified,deng2016similarity,yang2016consistent}. Very recently, DL-based SISR algorithms have demonstrated great superiority to reconstruction-based and other learning-based methods.

\subsection{Deep Learning}\label{ss:dl}
Deep learning is a branch of machine learning algorithms based on directly learning diverse representations of data~\cite{bengio2013representation}. In contrast to traditional task-specific learning algorithms that select useful handcrafted features with expert domain knowledge, deep learning algorithms aim to learn informative hierarchical representations automatically and then leverage them to achieve the final purpose, where the whole learning process can be seen as an entirety~\cite{song2013hierarchical}.  

Because of the high approximating capacity and hierarchical property of an artificial neural network (ANN), most modern deep learning models are based on ANNs~\cite{schmidhuber2015deep}. Early ANNs can be traced back to perceptron algorithms in the 1960s~\cite{rochester1956tests}. Then, in the 1980s, the multilayer perceptron could be trained with the backpropagation algorithm~\cite{rumelhart1986learning}, and the convolutional neural network (CNN)~\cite{lecun1989backpropagation} and recurrent neural network (RNN)~\cite{elman1990finding}, two representative derivatives of the traditional ANN, were introduced to the computer vision and speech recognition fields, respectively. Despite remarkable progress achieved by ANNs during that period, there were still many deficiencies handicapping ANNs from developing further~\cite{bengio1994learning,hochreiter2001gradient}. Thereafter, the rebirth of the modern ANN was marked by pretraining the deep neural network (DNN) with the restricted Boltzmann machine (RBM) proposed by Hinton in 2006~\cite{hinton2007learning}. Consequently, benefiting from the boom of computing power and the development of advanced algorithms, models based on the DNN have achieved remarkable performance in various supervised tasks~\cite{ciresan2011flexible,cirecsan2012multi,krizhevsky2012imagenet}. Meanwhile, DNN-based unsupervised algorithms such as the deep Boltzmann machine (DBM)~\cite{salakhutdinov2010efficient}, variational autoencoder (VAE)~\cite{kingma2013auto,rezende2014stochastic} and generative adversarial nets (GAN)~\cite{goodfellow2014generative} have attracted much attention owing to their potential to address challenging unlabeled data. Readers can refer to~\cite{goodfellow2016deep} for an extensive analysis of DL.

\section{Deep Architectures for SISR}\label{s:arch}

In this section, we mainly discuss the efficient architectures proposed for SISR in recent years. First, we set the network architecture of super-resolution CNN (SRCNN)~\cite{dong2014learning,dong2016image} as the benchmark. When we discuss each related architecture in detail, we focus on their universal parts that can apply to other tasks and their specific parts that characterize SISR properties. To meaningfully construct fair comparisons among different models, we will illustrate the importance of the training dataset and attempt to compare models with the same training dataset.

\begin{figure}[!t]
\centering
\includegraphics[scale=0.26]{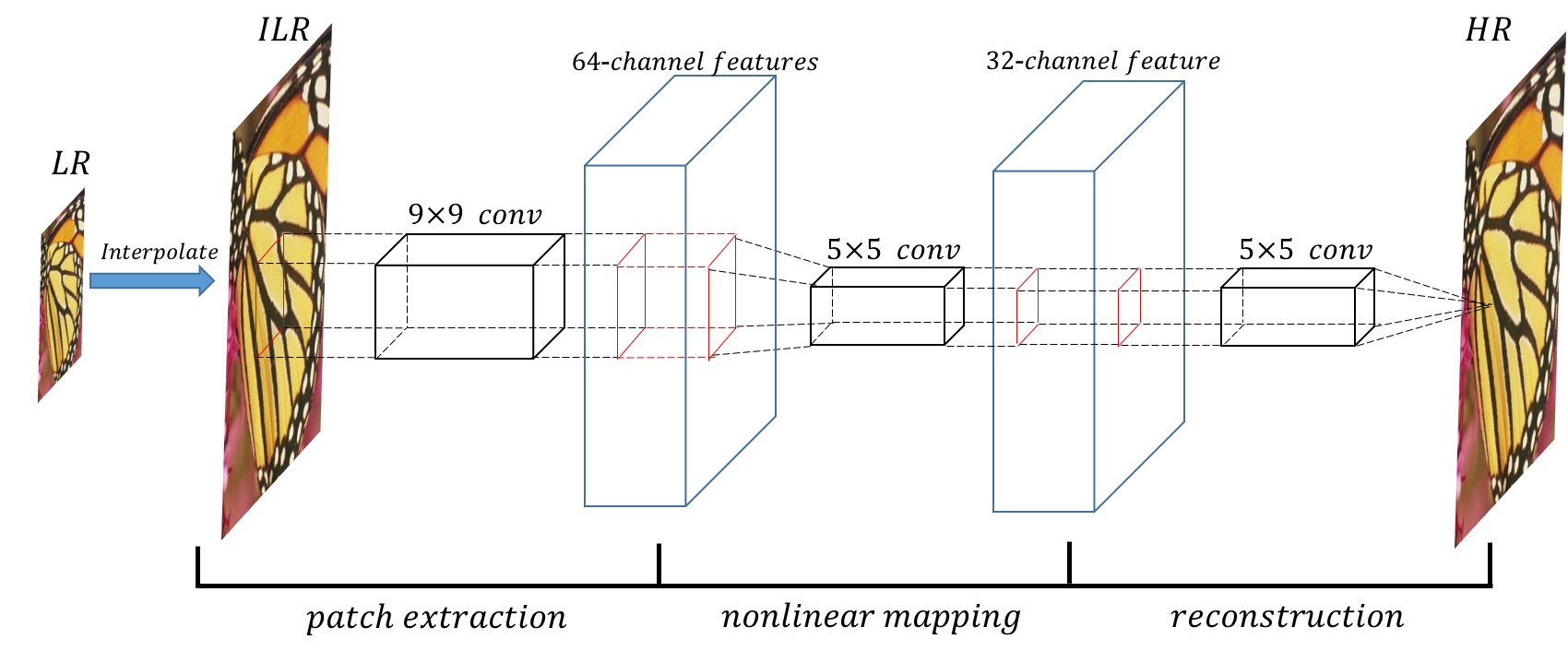}
\caption{Sketch of the SRCNN architecture.}
\label{SRCNN}
\end{figure}

\subsection{Benchmark of Deep Architecture for SISR}

We select the SRCNN architecture as the benchmark in this section. The overall architecture of SRCNN is shown in Fig.~\ref{SRCNN}. As established in many traditional methods, for simplicity, SRCNN only implements the luminance components for training. SRCNN is a three-layer CNN, where the filter sizes of each layer are $64 \times 1 \times 9 \times 9$, $32 \times 64 \times 5 \times 5$ and $1 \times 32 \times 5 \times 5$. The functions of these three nonlinear transformations are patch extraction, nonlinear mapping and reconstruction. The loss function for optimizing SRCNN is the mean square error (MSE), which will be discussed in the next section.

The formulation of SRCNN is relatively simple and can be envisioned as an ordinary CNN that approximates the complex mapping between the LR and HR spaces in an end-to-end manner. SRCNN reportedly demonstrated vast superiority over concurrent traditional methods, and we argue that its acclaim is owing to the CNN's strong capability of learning valid representations from big data in an end-to-end manner.

Despite the success of SRCNN, the following problems have inspired more effective architectures: 

1) The input of SRCNN is the bicubic LR, an approximation of HR. However, these interpolated inputs have three drawbacks: (a) detail-smoothing effects introduced by these inputs may lead to further wrong estimations of the image structure; (b) employing interpolated versions as input is very time-consuming; and (c) when the downsampling kernel is unknown, one specific interpolated input as a raw estimation is unreasonable. Therefore, the first question emerges: can we design CNN architectures that directly implement LR as input to address these problems?\footnote{Generally, the first problem can be grouped into the third problem below. Because the solutions to this problem form the basis of many other models, it is necessary to introduce this problem separately as the first drawback.}

2) The SRCNN is just a three-layer architecture. Can more complex CNN architectures (with different depths, widths and topologies) achieve better results? If yes, then how can we design such models of greater complexity?

3) The prior terms in the loss function that reflect properties of HR images are trivial. Can we integrate any property of the SISR process into the design of the CNN frame or other parts in the algorithms for SISR?  If yes, then can these deep architectures with SISR properties be more effective in addressing some challenging SISR problems, such as the large scale factor SISR and the unknown downsampling of SISR?

Based on some solutions to these three questions, recent studies on deep architectures for SISR will be discussed in Sections~\ref{sss:upsampling}, \ref{sss:deeper} and~\ref{sss:combining}.

\begin{figure}[!t]
\centering
\includegraphics[scale=0.2]{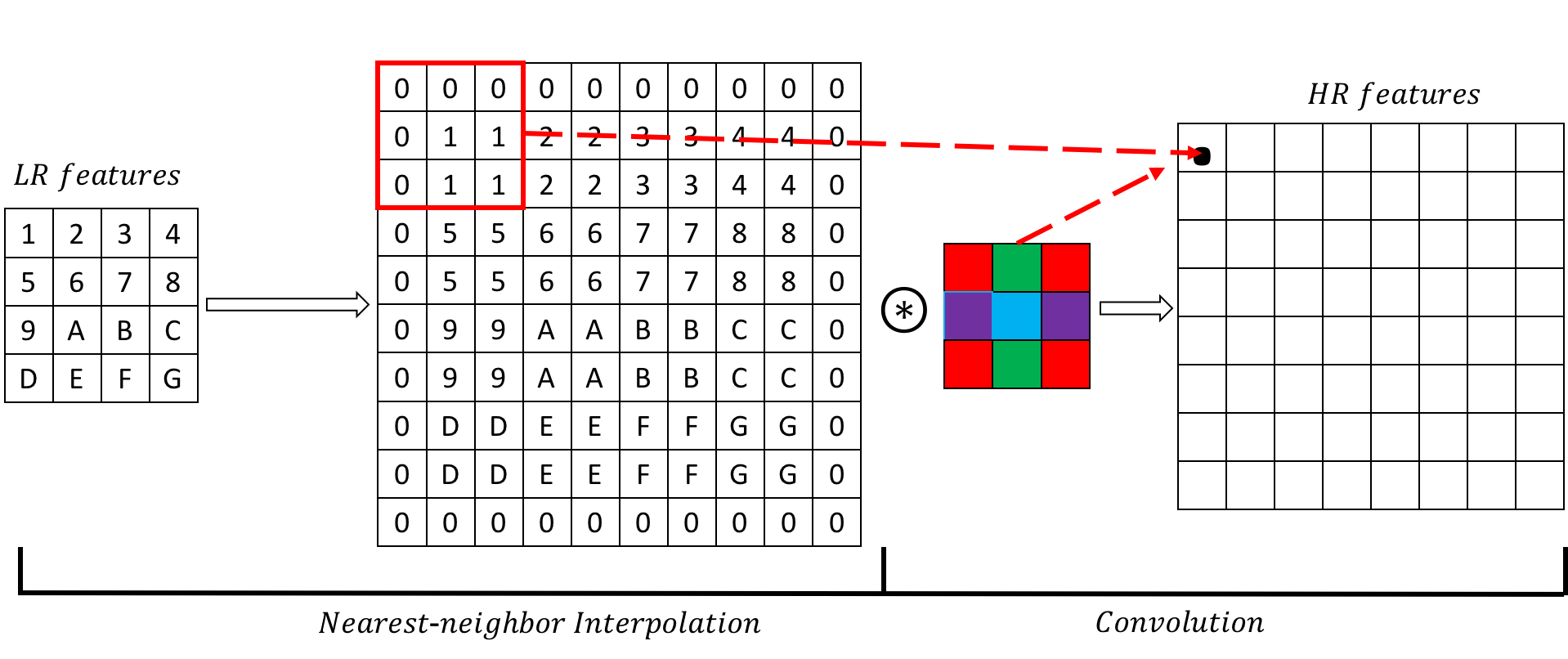}
\caption{Sketch of the deconvolution layer used in FSRCNN~\cite{dong2016image}, where $\circledast$ denotes the convolution operator.}
\label{FSRCNN}
\end{figure}

\begin{figure}[!t]
\centering
\includegraphics[scale=0.2]{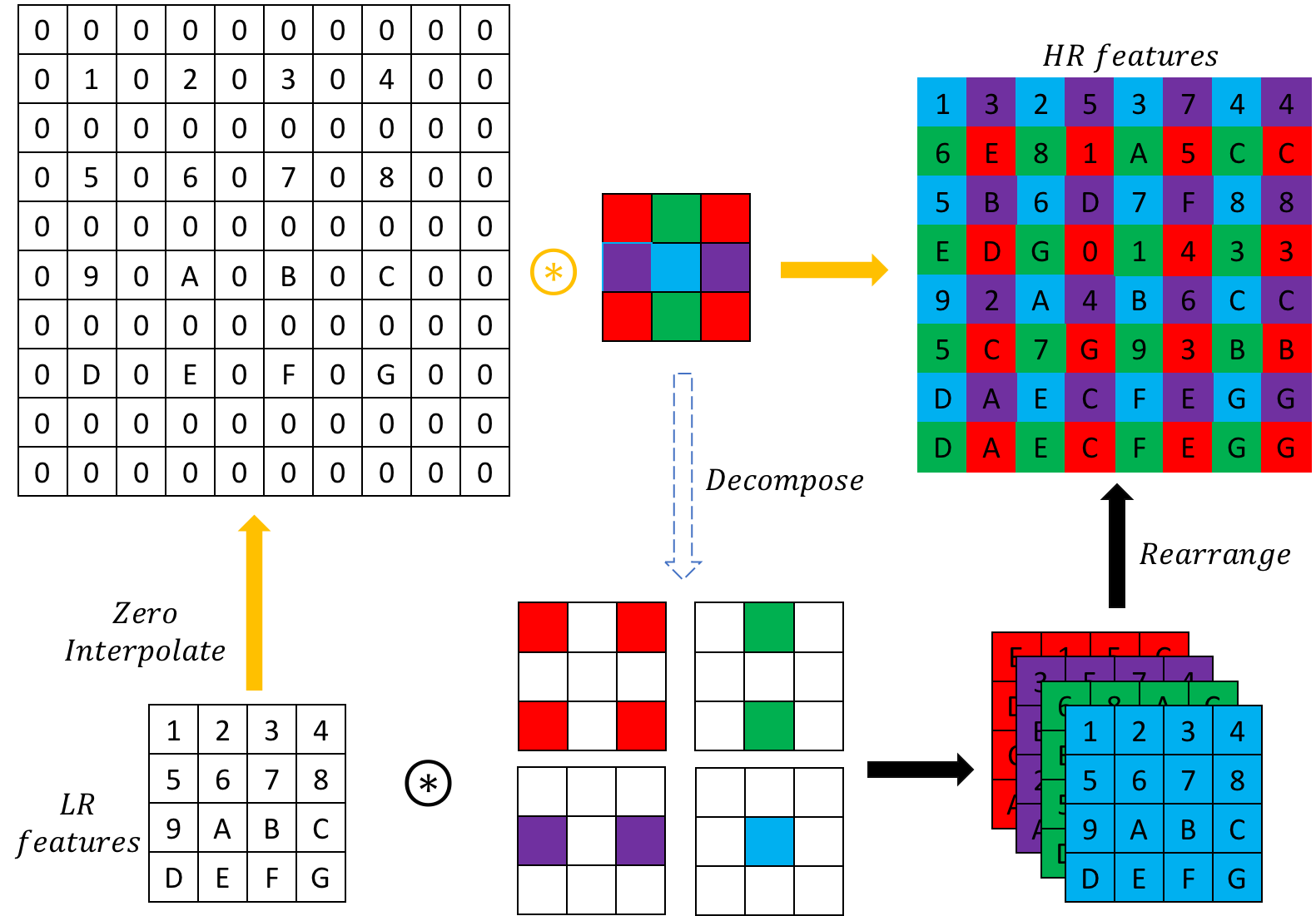}
\caption{Detailed sketch of ESPCN~\cite{shi2016real}. The top process with the yellow arrow depicts the ESPCN from the view of zero interpolation, while the bottom process with the black arrow is the original ESPCN; $\circledast$ denotes the convolution operator.}
\label{ESPCN}
\end{figure}

\subsection{State-of-the-Art Deep SISR Networks}

\subsubsection{Learning Effective Upsampling with CNN}\label{sss:upsampling}

One solution to the first question is to design a module in the CNN architecture that adaptively increases the resolution. Convolution with pooling and stride convolution are the common downsampling operators in the basic CNN architecture. Naturally, people can implement the upsampling operation, which is known as deconvolution~\cite{zeiler2011adaptive} or transposed convolution~\cite{dumoulin2016guide}. Given the upsampling factor, the deconvolution layer is composed of an arbitrary interpolation operator (usually, we choose the nearest neighbor interpolation for simplicity) and a following convolution operator with a stride of 1, as shown in Fig.~\ref{FSRCNN}. Readers should be aware that such deconvolution may not completely recover the information missing from convolution with pooling or stride convolution. Such a deconvolution layer has been successfully adopted in the context of network visualization~\cite{zeiler2014visualizing}, semantic segmentation~\cite{long2015fully} and generative modeling~\cite{radford2015unsupervised}. For a more detailed illustration of the deconvolution layer, readers can refer to~\cite{shi2016deconvolution}. To the best of our knowledge, FSRCNN~\cite{dong2016accelerating} is the first work using this normal deconvolution layer to reconstruct HR images from LR feature maps. As mentioned previously, the usage of the deconvolution layer has two main advantages: one is that a reduction in computation is achieved because we just need to increase resolution at the end of the network; the other is that when the downsampling kernel is unknown, many reports, e.g.,~\cite{efrat2013accurate}, have shown that when an inaccurate estimation is input, there are side effects on the final performance.

Although the normal deconvolution layer, which has already been involved in popular open source packages such as Caffe~\cite{jia2014caffe} and TensorFlow~\cite{abadi2016tensorflow}, offers a reasonably good solution to the first question, there is still an underlying problem: when we use the nearest neighbor interpolation, the points in the upsampled features are repeated several times in each direction. This configuration of the upsampled pixels is redundant. To circumvent this problem, Shi \emph{et~al.} proposed an efficient subpixel convolution layer in~\cite{shi2016real}, known as ESPCN; the structure of ESPCN is shown in Fig.~\ref{ESPCN}. Rather than increasing resolution by explicitly enlarging feature maps as the deconvolution layer does, ESPCN expands the channels of the output features for storing the extra points to increase resolution and then rearranges these points to obtain the HR output through a specific mapping criterion. As the expansion is carried out in the channel dimension, a smaller kernel size is sufficient. \cite{shi2016deconvolution} further shows that when the ordinary but redundant nearest neighbor interpolation is replaced with the interpolation that pads the subpixels with zeroes, the deconvolution layer can be simplified into the subpixel convolution in ESPCN. Obviously, compared with the nearest neighbor interpolation, this interpolation is more efficient, which can also verify the effectiveness of ESPCN.

\subsubsection{The Deeper, The Better}\label{sss:deeper}
\begin{figure*}
    \centering
    \subfloat[VDSR]{
       \includegraphics[scale=0.3]{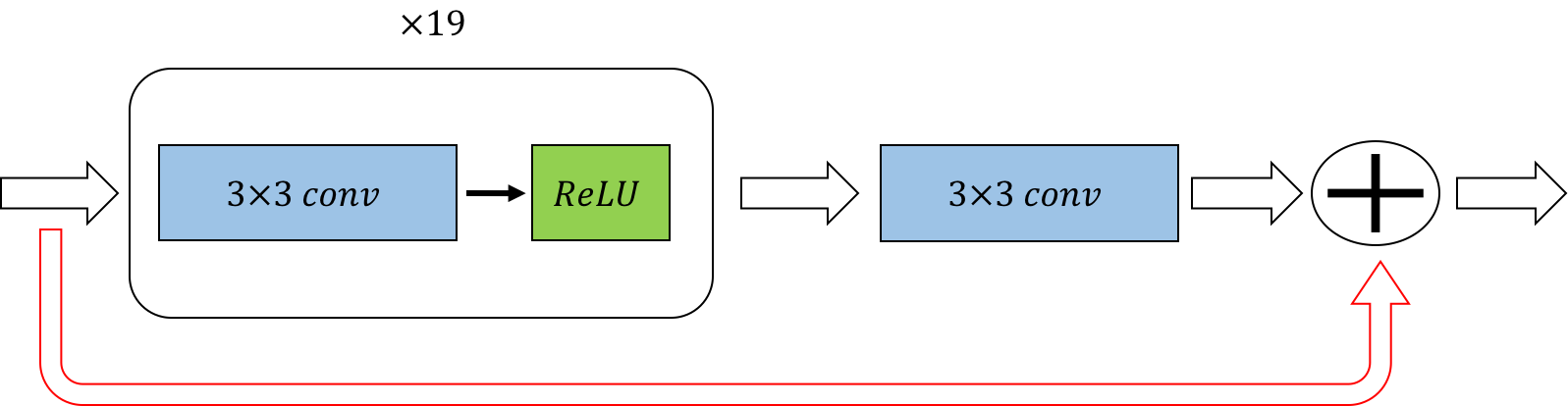}}
      \label{VDSR}\hfill
    \subfloat[DRCN]{
    	\includegraphics[scale=0.3]{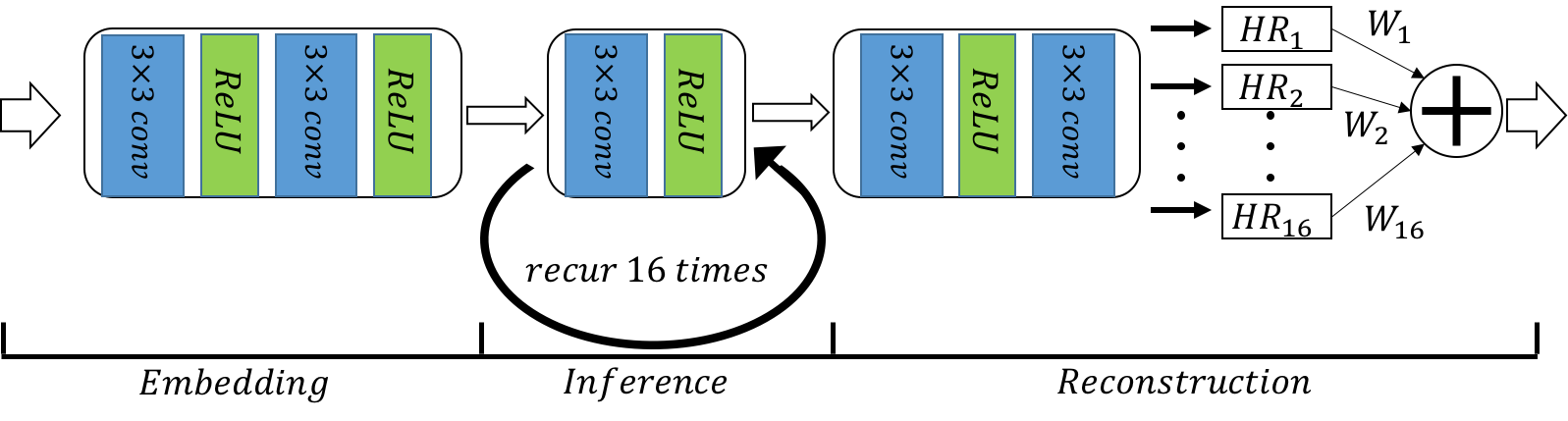}}
      \label{DRCN}\\
    \subfloat[SRResNet]{
    	\includegraphics[scale=0.3]{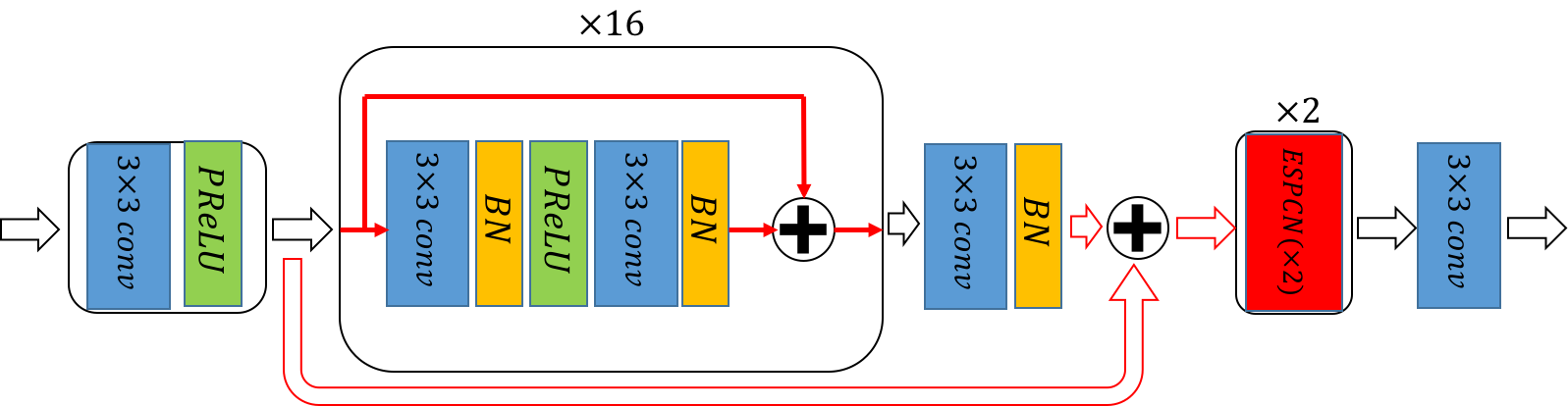}}
      \label{SRResNet}\hfill
    \subfloat[DRRN]{
    	\includegraphics[scale=0.3]{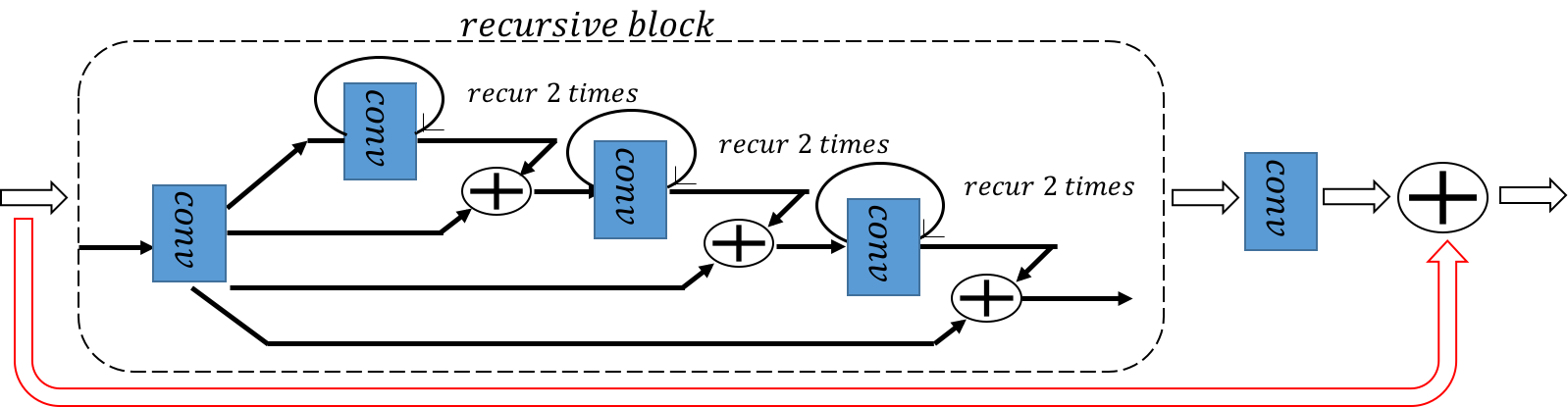}}
      \label{DRRN}\\
    \subfloat[EDSR]{
    	\includegraphics[scale=0.3]{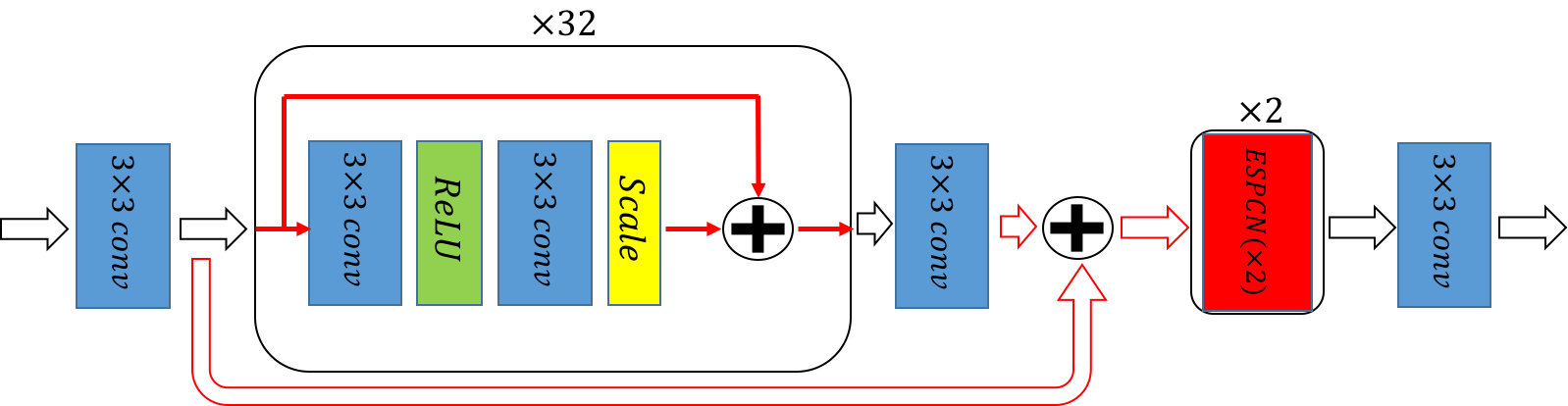}}
      \label{EDSR}\hfill
    \subfloat[DenseSR]{
    	\includegraphics[scale=0.3]{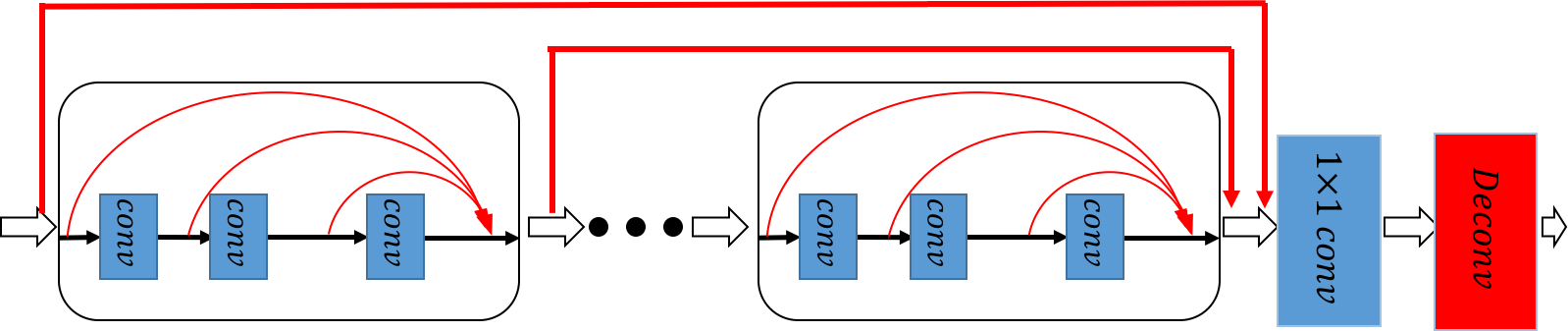}}
      \label{DenseSR}\\
    \subfloat[MDSR]{
    	\includegraphics[scale=0.25]{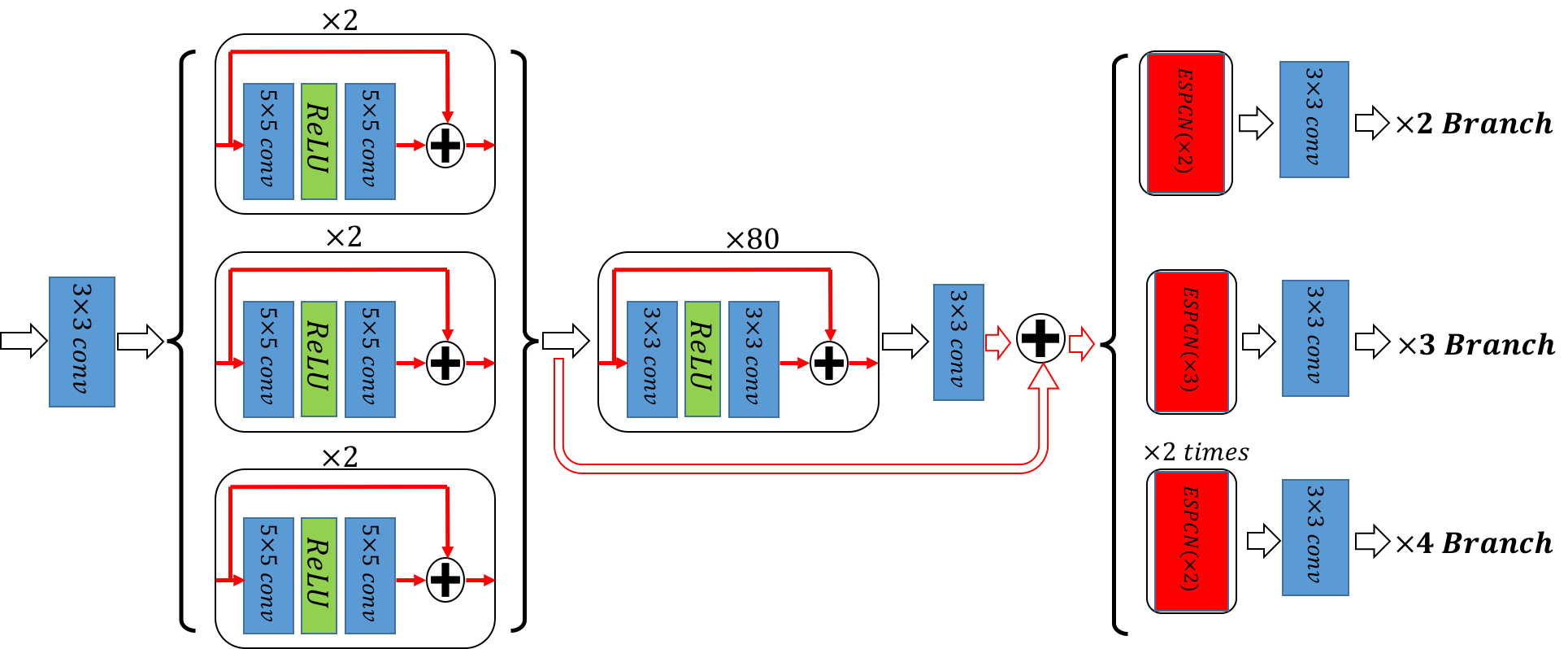}}
      \label{MDSR}\hfill
    \subfloat[MemNet]{
    	\includegraphics[scale=0.25]{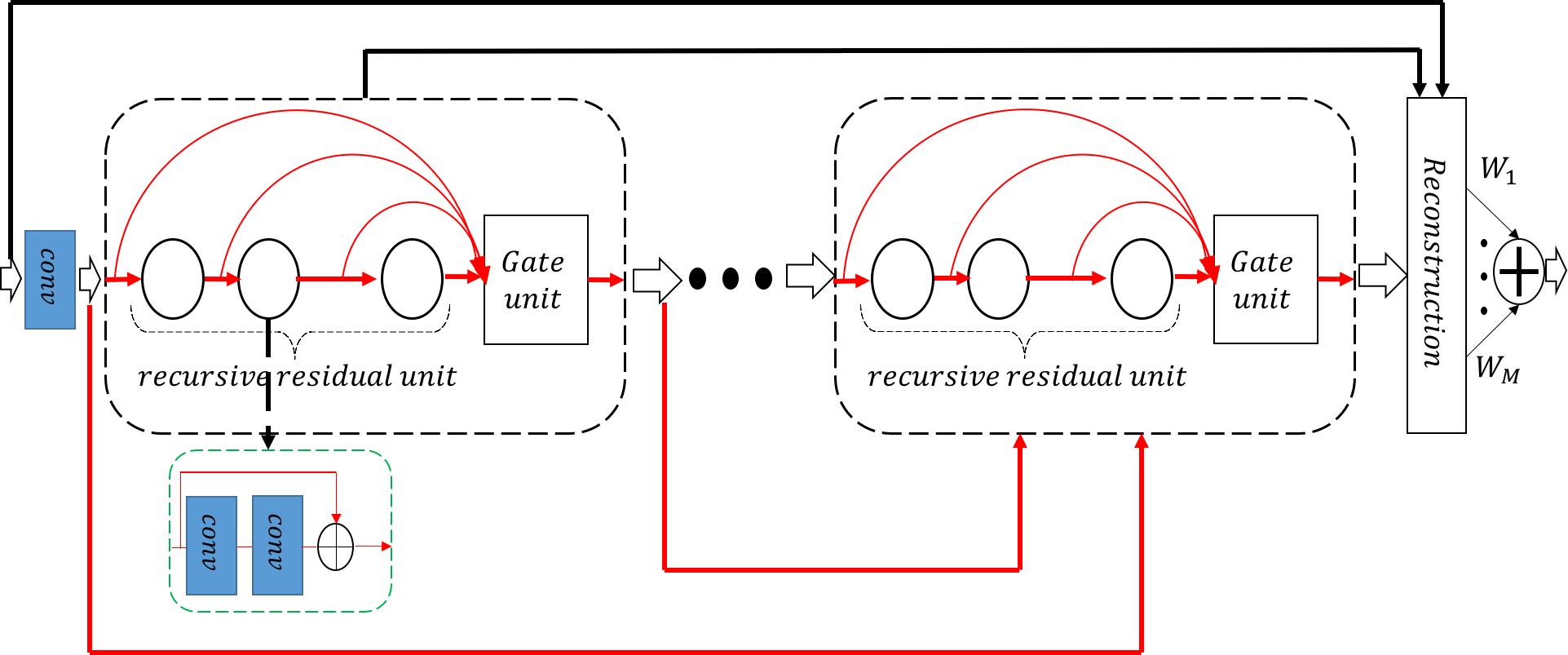}}
      \label{MemNet}\\
    \caption{Sketch of several deep architectures for SISR.}
    \label{comparison}
\end{figure*}

In the DL research, there is theoretical work~\cite{montufar2014number} showing that the solution space of a DNN can be expanded by increasing its depth or its width. In some situations, to attain more hierarchical representations more effectively, many works mainly focus on improvements acquired by increasing the depth. Recently, various DL-based applications have also demonstrated the great power of very deep neural networks despite many training difficulties. VDSR~\cite{kim2016accurate} is the first very deep model used in SISR. As shown in Fig.~\ref{comparison}(a), VDSR is a 20-layer VGG-net~\cite{simonyan2014very}. The VGG architecture sets all kernel sizes as $3 \times 3$ (the kernel size is usually odd and takes the increase in the receptive field into account, and $3 \times 3$ is the smallest kernel size). To train this deep model, the authors used a relatively high initial learning rate to accelerate convergence and used gradient clipping to prevent the annoying gradient explosion problem. 

In addition to the innovative architecture, VDSR has made two more contributions. The first one is that a single model is used for multiple scales since the SISR processes with different scale factors have a strong relationship with each other. This fact is the basis of many traditional SISR methods. Similar to SRCNN, VDSR takes the bicubic of LR as input. During training, VDSR puts the bicubics of LR of different scale factors together for training. For larger scale factors ($\times 3, \times 4$), the mapping for a smaller scale factor ($\times 2$) may also be informative. 
The second contribution is the residual learning. Unlike the direct mapping from the bicubic version to HR, VDSR uses deep CNN to learn the mapping from the bicubic to the residual between the bicubic and HR. The authors argued that residual learning could improve performance and accelerate convergence. 

The convolution kernels in the nonlinear mapping part of VDSR are very similar, and in order to reduce parameters, Kim \emph{et~al.} further proposed DRCN~\cite{kim2016deeply}, which utilizes the same convolution kernel in the nonlinear mapping part 16 times, as shown in Fig.~\ref{comparison}(b). To overcome the difficulties of training a deep recursive CNN, a multisupervised strategy is applied, and the final result can be regarded as the fusion of 16 intermediate results. The coefficients for fusion are a list of trainable positive scalars with the summation of 1. As they showed, DRCN and VDSR have a quite similar performance. 

Here, we believe that it is necessary to emphasize the importance of the multisupervised training in DRCN. This strategy not only creates short paths through which the gradients can flow more smoothly during backpropagation but also guides all the intermediate representations to reconstruct raw HR outputs. Finally, fusing all these raw HR outputs produces a wonderful result. However, for fusion, this strategy has two flaws: 1) once the weight scalars are determined in the training process, they will not change with different inputs; and 2) using a single scalar to weight HR outputs does not take pixelwise differences into consideration, that is, it would be better to weight different parts distinguishingly in an adaptive way.

It is hard to go deeper with a plain architecture such as VGG-net. Various deep models based on skip-connections can be extremely deep and have achieved state-of-the-art performance in many tasks. Among them, ResNet~\cite{he2016deep,he2016identity}, proposed by He \emph{et~al.}, is the most representative model. Readers can refer to~\cite{veit2016residual,balduzzi2017shattered} for further discussions on why ResNet works well. In~\cite{ledig2017photo}, the authors proposed SRResNet, which is composed of 16 residual units (a residual unit consists of two nonlinear convolutions with residual learning). In each unit, batch normalization (BN)~\cite{ioffe2015batch} is used to stabilize the training process. The overall architecture of SRResNet is shown in Fig.~\ref{comparison}(c). Based on the original residual unit in~\cite{he2016identity}, Tai \emph{et~al.} proposed DRRN~\cite{tai2017image}, in which basic residual units are rearranged in a recursive topology to form a recursive block, as shown in Fig.~\ref{comparison}(d). Then, to accommodate parameter reduction, each block shares the same parameters and is reused recursively, such as in the single recursive convolution kernel in DRCN.  

EDSR~\cite{lim2017enhanced} was proposed by Lee \emph{et~al.} and has currently achieved state-of-the-art performance. EDSR has mainly made three improvements on the overall frame: 1) Compared with the residual unit used in previous work, EDSR removes the usage of BN, as shown in Fig.~\ref{comparison}(e). The original ResNet with BN was designed for classification, where inner representations are highly abstract, and these representations can be insensitive to the shift introduced by BN. Regarding image-to-image tasks such as SISR, since the input and output are strongly related, if the convergence of the network is not a problem, then such a shift may harm the final performance. 2) Except for regular depth increasing, EDSR also increases the number of output features of each layer on a large scale. To relinquish the difficulties of training such a wide ResNet, the residual scaling trick proposed in ~\cite{szegedy2017inception} is employed.
3) Additionally, inspired by the fact that the SISR processes with different scale factors have strong relationships with each other, when training the models for $\times 3$ and $\times 4$ scales, the authors of~\cite{lim2017enhanced} initialized the parameters with the pretrained $\times 2$ network. This pretraining strategy accelerates the training and improves the final performance.

The effectiveness of the pretraining strategy in EDSR implies that models for different scales may share many intermediate representations. To explore this idea further, similar to building a multiscale architecture as VDSR does on the condition of bicubic input, the authors of EDSR proposed MDSR to achieve the multiscale architecture, as shown in Fig.~\ref{comparison}(g). In MDSR, the convolution kernels for nonlinear mapping are shared across different scales, where only the front convolution kernels for extracting features and the final subpixel upsampling convolution are different. At each update during training MDSR, minibatches for $\times 2$, $\times 3$ and $\times 4$ are randomly chosen, and only the corresponding parts of MDSR are updated. 

In addition to ResNet, DenseNet~\cite{huang2017densely} is another effective architecture based on skip connections. In DenseNet, each layer is connected with all the preceding representations, and the bottleneck layers are used in units and blocks to reduce the parameter amounts. In~\cite{chen2017dual}, the authors pointed out that ResNet enables feature re-usage while DenseNet enables new feature exploration. Based on the basic DenseNet, SRDenseNet~\cite{tong2017image}, as shown in Fig.~\ref{comparison}(f), further concatenates all the features from different blocks before the deconvolution layer, which is shown to be effective in improving performance. MemNet~\cite{tai2017memnet}, proposed by Tai \emph{et~al.}, uses the residual unit recursively to replace the normal convolution in the block of the basic DenseNet and adds dense connections among different blocks, as shown in Fig.~\ref{comparison}(h). The authors explained that the local connections in the same block resemble the short-term memory and the connections with previous blocks resemble the long-term memory~\cite{hochreiter1997long}. Recently, RDN~\cite{zhang2018residual} was proposed by Zhang \emph{et~al.} and uses a similar structure. In an RDN block, basic convolution units are densely connected similar to DenseNet, and at the end of an RDN block, a bottleneck layer is used, following with the residual learning across the whole block. Before entering the reconstruction part, features from all previous blocks are fused by the dense connection and residual learning. 

\subsubsection{Combining Properties of the SISR Process with the Design of the CNN Frame}\label{sss:combining}
In this subsection, we discuss some deep frames whose architectures or procedures are inspired by some representative methods for SISR. Compared with the abovementioned NN-oriented methods, these methods can be better interpreted, and they sometimes are more sophisticated in addressing certain challenging cases for SISR.

\textbf{Combining sparse coding with deep NN:} The sparse prior in nature images and the relationships between the HR and LR spaces rooted from this prior were widely used for their great performance and theoretical support. SCN~\cite{wang2015deep} was proposed by Wang \emph{et~al.} and uses the learned iterative shrinkage and thresholding algorithm (LISTA)~\cite{gregor2010learning}, which produces an approximate estimation of sparse coding based on NN, to solve the time-consuming inference in traditional sparse coding SISR. They further introduced a cascaded version (CSCN)~\cite{liu2016robust} that employs multiple SCNs. Previous works such as SRCNN tried to explain general CNN architectures with the sparse coding theory, which from today's view may be somewhat unconvincing. SCN combines these two important concepts innovatively and gains both quantitative and qualitative improvements. 

\textbf{Learning to ensemble by NN: }Different models specialize in different image patterns of SISR. From the perspective of ensemble learning, a better result can be acquired by adaptively fusing various models with different purposes at the pixel level. Motivated by this idea, MSCN was proposed by Liu \emph{et~al.}~\cite{liu2016learning} by developing an extra module in the form of a CNN, taking the LR as input and outputting several tensors with the same shape as the HR. These tensors can be viewed as adaptive elementwise weights for each raw HR output. By selecting NNs as the raw SR inference modules, the raw estimating parts and the fusing part can be optimized jointly. However, in MSCN, the summation of coefficients at each pixel is not 1, which may be slightly incongruous.

\textbf{Deep architectures with progressive methodology:} Increasing SISR performance progressively has been extensively studied previously, and many recent DL-based approaches also exploit it from various perspectives. Here, we mainly discuss three novel works within this scope: DEGREE~\cite{yang2017deep}, combining the progressive property of ResNet with traditional subband reconstruction; LapSRN~\cite{lai2017deep}, generating SR of different scales progressively; and PixelSR~\cite{dahl2017pixel}, leveraging conditional autoregressive models to generate SR pixel-by-pixel. 

Compared with other deep architectures, ResNet is intriguing for its progressive properties. Taking SRResNet for example, one can observe that directly sending the representations produced by intermediate residual blocks to the final reconstruction part will also yield a quite good raw HR estimator. The deeper these representations are, the better the results that can be obtained. A similar phenomenon of ResNet applied in recognition is reported in~\cite{veit2016residual}. DEGREE, proposed by Yang \emph{et~al.}, combines this progressive property of ResNet with the subband reconstruction of traditional SR methods~\cite{singh2014super}. The residues learned in each residual block can be used to reconstruct high-frequency details, resembling the signals from a certain high-frequency band. To simulate subband reconstruction, a recursive residual block is used. Compared with the traditional supervised subband recovery methods that need to obtain subband ground truth by diverse filters, this simulation with recursive ResNet avoids explicitly estimating intermediate subband components, benefiting from the end-to-end representation learning. 

\begin{figure}[!t]
\centering
\includegraphics[scale=0.25]{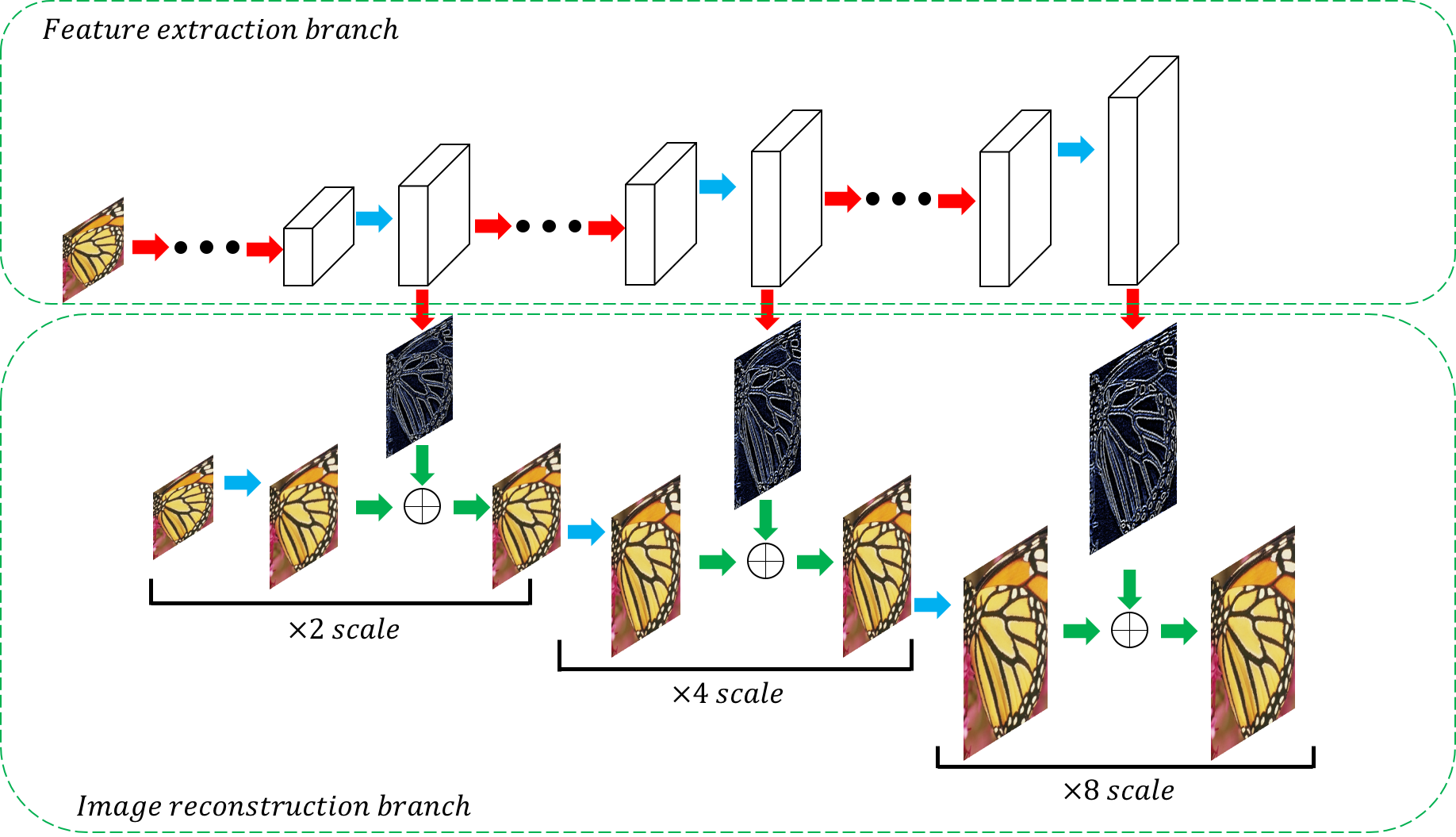}
\caption{LapSRN architecture. Red arrows indicate the convolutional layer; blue arrows indicate transposed convolutions (upsampling); green arrows denote elementwise addition operators.}
\label{LapSRN}
\end{figure}

As mentioned above, models for small scale factors can be used for a raw estimator of a large scale SISR. In the SISR community, SISR under large scale factors (\emph{e.g.,}$\times$8) has been a challenging problem for a long time. In such situations, plausible priors are imposed to restrict the solution space. A straightforward way to address this is to gradually increase resolution by adding extra supervision on the auxiliary SISR process of the small scale. Based on this heuristic prior, LapSRN, proposed by Lai \emph{et~al.}, uses the Laplacian pyramid structure to reconstruct HR outputs. LapSRN has two branches: the feature extraction branch and the image reconstruction branch, as shown in Fig.~\ref{LapSRN}. At each scale, the image reconstruction branch estimates a raw HR output of the present stage, and the feature extraction branch outputs a residue between the raw estimator and the corresponding ground truth as well as extracts useful representations for the next stage. 

\begin{figure}[!t]
\centering
\includegraphics[scale=0.3]{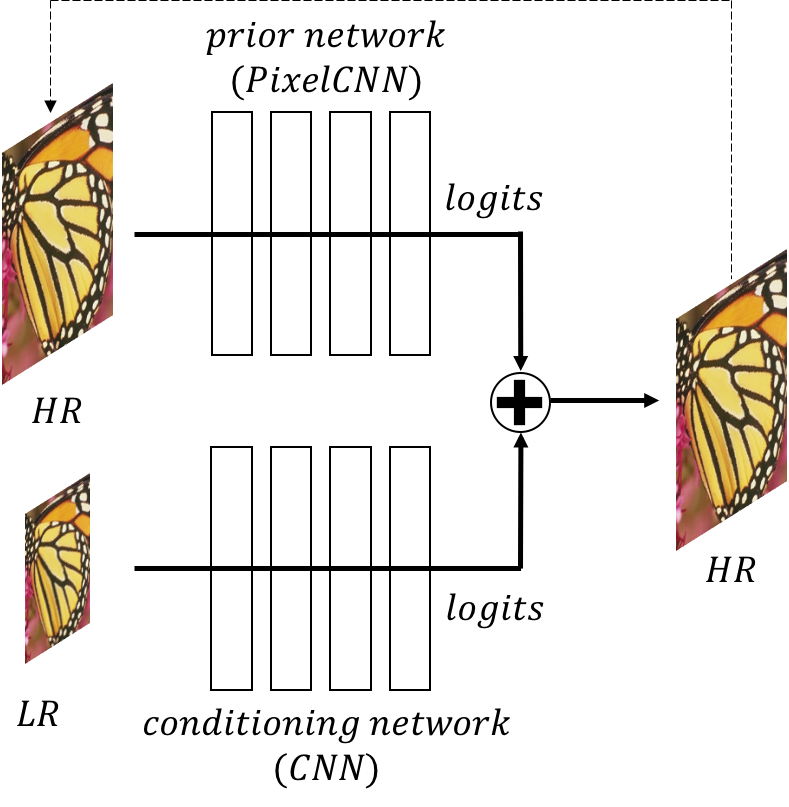}
\caption{Sketch of the pixel recursive SR architecture.}
\label{pixelSR}
\end{figure}

When faced with large scale factors with a severe loss of necessary details, some researchers suggest that synthesizing rational details can achieve better results. In this situation, deep generative models, which will be discussed in the next sections, could be good choices. Compared with the traditional independent point estimation of the lost information, conditional autoregressive generative models using conditional maximum likelihood estimation in directional graphical models gradually generate high-resolution images based on the previously generated pixels. PixelRNN~\cite{oord2016pixel} and PixelCNN~\cite{van2016conditional} are recent representative autoregressive generative models. The current pixel in PixelRNN and PixelCNN is explicitly dependent on the left and top pixels that have already been generated. To implement such operations, novel network architectures are elaborated. PixelSR was proposed by Dahl \emph{et~al.} and first applies conditional PixelCNN to SISR. The overall architecture is shown in Fig.~\ref{pixelSR}. The conditioning CNN takes LR as input, which provides LR-conditional information to the whole model, and the PixelCNN part is the autoregressive inference part. The current pixel is determined by these two parts together using the current softmax probability:
\begin{equation}
\begin{split}
P(y_{i}|x, y_{<i})={\rm softmax} (A_{i}(x)+B_{i}(y_{<i})),
\end{split}
\label{pixelcnn_1}
\end{equation}
where $x$ is the LR input, $y_{i}$ is the current HR pixel to be generated, $y_{<i}$ are the generated pixels, $A_{i}(\cdot)$ denotes the conditioning network predicting a vector of logit values corresponding to the possible values, and $B_{i}(\cdot)$ denotes the prior network predicting a vector of logit values of the $i$th output pixel. Pixels with the highest probability are taken as the final output pixel. 

Similarly, the whole network is optimized by minimizing cross-entropy loss (maximizing the corresponding log-likelihood) between the model's prediction and the discrete ground-truth labels.

\begin{figure}[!t]
\centering
\includegraphics[scale=0.3]{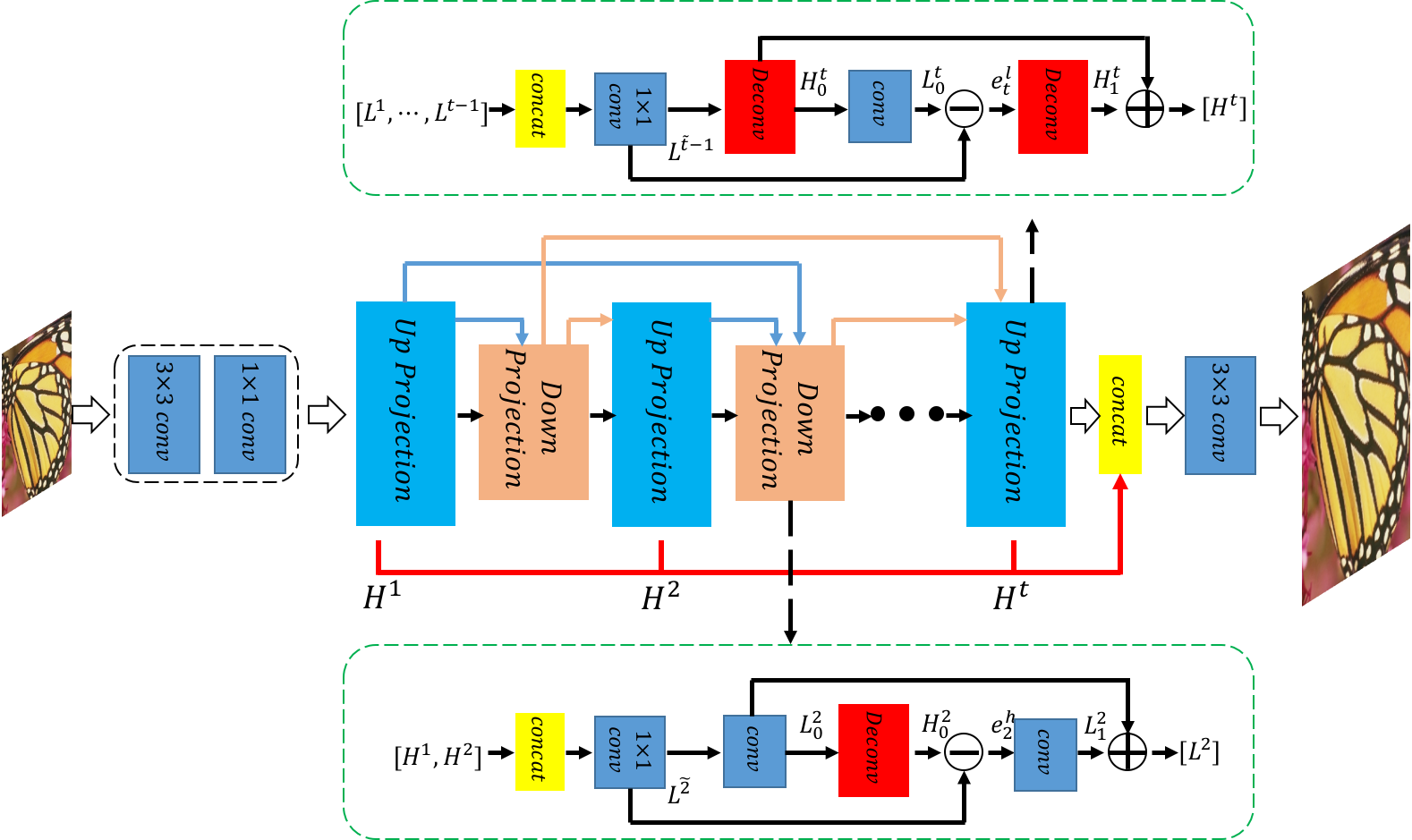}
\caption{Sketch of the DBPN architecture.}
\label{DBPN}
\end{figure}

\textbf{Deep architectures with backprojection: }Iterative backprojection~\cite{irani1991improving} is an early SR algorithm that iteratively computes the reconstruction error and then feeds it back to tune the HR results. Recently, DBPN~\cite{haris2018deep}, proposed by Haris \emph{et~al.}, uses deep architectures to simulate iterative backprojection and further improves performance with dense connections~\cite{huang2017densely}, which is shown to achieve wonderful performance in the $\times8$ scale. As shown in Fig.~\ref{DBPN}, the dense connection and $1\times1$ convolution for reducing the dimension is first applied across different up-projection (down-projection) units; next, in the $t$th up-projection unit, the current LR feature input $L^{\widetilde{t}-1}$ is first deconvoluted to obtain a raw HR feature $H_{0}^{t}$, and $H_{0}^{t}$ is backprojected to the LR feature $L_{0}^{t}$. The residue between two LR features $e_{t}^{l}=L^{t-1}-L_{0}^{t-1}$ is then deconvoluted and added to $H_{0}^{t}$ to obtain a finer HR feature $H^{t}$. The down-projection unit is defined very similarly in an inverse way. 

\textbf{Usage of additional information from LR: }Although modern deep NNs are skillful in extracting various ranges of useful representations in end-to-end manners, in some cases, it is still helpful to select some information to process explicitly. For example, the DEGREE~\cite{yang2017deep} takes the edge map of LR as another input. Recent studies tend to use more complex information of LR directly, two examples of which are the following: SFT-GAN~\cite{wang2018recovering}, with extra semantic information of LR for better perceptual quality, and SRMD~\cite{zhang2018learning}, incorporating degradation into input for multiple degradations. 

\cite{timofte2016semantic} reported that using a semantic prior helps improve the performance of many SISR algorithms. Leveraging powerful deep architectures recently designed for segmentation, Wang \emph{et~al.}~\cite{wang2018recovering} used semantic segmentation maps of interpreted LR as additional input and deliberated the spatial feature transformation (SFT) layer to handle them. With this extra information from high-level tasks, the proposed work is more skilled in generating textual details. 

To take degradations of different LRs into account, SRMD first applied a parametric zero-mean anisotropic Gaussian kernel to stand for the blur kernel and the additive white Gaussian noise with hyperparameter ${\rho}^2$ to represent noise. Then, a simple regression is used to obtain its covariance matrix. These sufficient statistics are dimensionally stretched to concatenate with LR in the channel dimension, and with such input, a deep model is trained. Notably, when SRMD is tested with real images, the needed parameters on the degradation level are obtained by grid search.

\textbf{Reconstruction-based frameworks based on priors offered by deep NN: }Sophisticated priors are of key points for efficient reconstruction-based SISR algorithms to address different cases flexibly. Recent works showed that deep NNs could provide well-performing priors mainly from two perspectives: priors in the deep NN learn from data in advance within a plug-and-play approach and direct reconstruction of output, leveraging intriguing but still unclear priors of deep architectures themselves. 

Given the degraded version $y$, the reconstruction-based algorithms aim to obtain the desired result $\hat{x}$ by solving 
\begin{equation}
    \begin{split}
        \hat{x}=\argmin{||Hx-y||_{2}^{2}+R(x)},
    \end{split}
    \label{MAP}
\end{equation}
where $H$ is the degradation matrix and $R(x)$ is regularization, also called a prior from the Bayesian view. \cite{venkatakrishnan2013plug} split (\ref{MAP}) into a data part and a prior part with variable splitting techniques and then replaced the prior part with efficient denoising algorithms. Regarding different degradation cases, one only needs to change denoising algorithms for the prior part, behaving in so-called plug-and-play manners. Recent works \cite{meinhardt2017learning,zhang2017learning,tirer2019image} use deep discriminatively trained NNs under different noise levels as denoisers in various inverse problems, and IRCNN \cite{zhang2017learning} is the first one among them to address SISR. In IRCNN, they first trained a series of CNN-based denoisers with different noise levels, and took backprojection as the reconstruction part. The LR is first preceded by several backprojection iterations and then denoised by CNN denoisers with decreasing noise levels along with backprojection. The iteration number is empirically set to 30. In IRCNN, the authors use deep networks to learn a set of image priors and then plug the priors into the reconstruction framework; the experimental results in these cases are better than the contemporary methods that only employ example-based training. 

\begin{figure}[!t]
\centering
\includegraphics[scale=0.4]{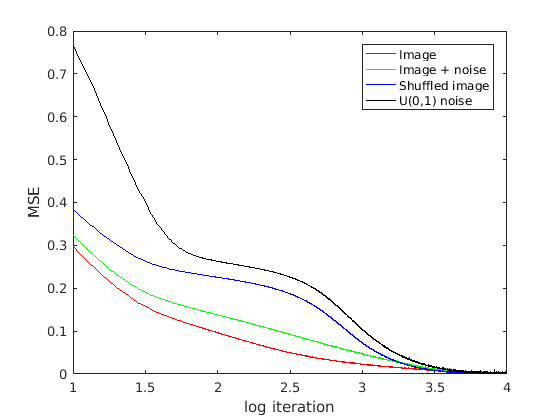}
\caption{Learning curves for the reconstruction of different kinds of images. We re-implement the experiment in \cite{ulyanov2017deep} with the image `butterfly' in Set5.}
\label{deep_prior}
\end{figure}

Recently, Ulyanov \emph{et~al.} showed in~\cite{ulyanov2017deep} that the structure of deep neural networks could capture a considerable amount of low-level image statistical priors. They reported that when neural networks are used to fit images of different statistical properties, the convergence speed for different kinds of images can also be different. As shown in Fig.~\ref{deep_prior}, natural-looking images, whose different parts are highly relevant, will converge much faster. In contrast, images such as noises and shuffled images, which have little inner relationship, tend to converge more slowly. Many inverse problems such as denoising and super-resolution are modeled as the pixel-wise summation of the original image and the independent additive noises. Based on the observed prior, when used to fit these degraded images, the neural networks tend to fit the natural-looking images first, which can be used to retain the natural-looking parts as well as to filter the noisy ones. To illustrate the effectiveness of the proposed prior for SISR, only given the LR $x_{0}$, the authors took a fixed random vector $z$ as input to fit the HR $x$ with a randomly initialized DNN $f_{\theta}$ by optimizing 
\begin{equation}
\begin{split}
\min \limits_{\theta} ||d(f_{\theta}(z))-x_{0}||_{2}^{2},
\end{split}
\label{prior_recon}
\end{equation}
where $d(\cdot)$ is a common differentiable downsampling operator. The optimization is terminated in advance for only filtering noisy parts. Although these totally unsupervised methods are outperformed by other supervised learning methods, they perform considerably better than some other naive methods.

\textbf{Deep architectures with internal examples: }Internal-example SISR algorithms are based on the recurrence of small pieces of information across different scales of a single image, which are shown to be better at addressing specific details rarely existing in other external images~\cite{zhang2013single}. ZSSR~\cite{shocher2017zero}, proposed by Shocher \emph{et~al.}, is the first literature combining deep architectures with internal-example learning. In ZSSR, other than the image for testing, no extra images are needed, and all the patches for training are taken from different degraded pairs of the test image. As demonstrated in~\cite{zontak2011internal}, the visual entropy inside a single image is much smaller than the large training dataset collected from wide ranges, so unlike external-example SISR algorithms, a very small CNN is sufficient. As we mentioned previously for VDSR, the training data for a small-scale model can also be useful for training large-scale models. Additionally, based on this trick, ZSSR can be more robust by collecting more internal training pairs with small scale factors for training large-scale models. However, this approach will increase runtime immensely. Notably, when combined with the kernel estimation algorithms mentioned in \cite{michaeli2013nonparametric}, ZSSR performs quite well with the unknown degradation kernels.

Recently, Tirer \emph{et~al.} argued that degradation in LR decreases the performance of internal-example algorithms~\cite{tirer2018super}. Therefore, they proposed to use reconstruction-based deep frame IDBP~\cite{tirer2019image} to obtain an initial SR result and then conduct internal-example-based network training similar to ZSSR. This method was believed to combine two successful techniques that address the mismatch between training and test, and it has achieved robust performance in these cases.

\subsection{Comparisons among Different Models and Discussion}

In this section, we will summarize recent progress in deep architectures for SISR from two perspectives: quantitative comparisons for those trained by specific blurring, and comparisons on those models for handling nonspecific blurring.

For the first part, quantitative criteria mainly include the following:

\textbf{1) PSNR/SSIM~\cite{wang2004image} for measuring reconstruction quality:} Given two images $I$ and $\hat I$ both with $N$ pixels, the MSE and peak signal-to-noise ratio (PSNR) are defined as
\begin{equation}
    \begin{split}
        MSE = \frac{1}{N}{||I - {\hat I}||}_{F}^{2},
    \end{split}
    \label{MSE_1}
\end{equation}
\begin{equation}
    \begin{split}
        PNSR = 10\log_{10}^{(\frac{L^2}{MSE})},
    \end{split}
    \label{PSNR}
\end{equation}
where $||\cdot||_{F}^2$ is the Frobenius norm and L is usually 255. The structural similarity index (SSIM) is defined as 
\begin{equation}
    \begin{split}
        SSIM(I,{\hat I})=\frac{2\mu_{I}\mu_{\hat I}+k_1}{\mu_{I}^{2}+\mu_{\hat I}^{2}+k_1}\cdot \frac{\sigma_{I\hat I}+k_2}{\sigma_{I}^2+\sigma_{\hat I}^2+k_2},
    \end{split}
    \label{SSIM}
\end{equation}
where $\mu_I$ and $\sigma_I^2$ is the mean and variance of $I$, $\sigma_{I {\hat I}}$ is the covariance between $I$ and $\hat I$, and $k_1$ and $k_2$ are constant relaxation terms.

\textbf{2) Number of parameters of NN for measuring storage efficiency (Params).}

\textbf{3) Number of composite multiply-accumulate operations for measuring computational efficiency (Mult\&Adds):} Since operations in NNs for SISR are mainly multiplications with additions, we use Mult\&Adds in CARN~\cite{ahn2018fast} to measure computation, assuming that the desired SR is 720p.

Notably, it has been shown in~\cite{dong2016image} and~\cite{shi2016real} that the training datasets have a great influence on the final performance, and usually, more abundant training data will lead to better results. Generally, these models are trained via three main datasets: 1) 91 images from~\cite{yang2010image} and 200 images from~\cite{martin2001database}, called the 291 dataset (some models only use 91 images); 2) images derived from ImageNet~\cite{deng2009imagenet} randomly; and 3) the newly published DIV2K dataset~\cite{agustsson2017ntire}. In addition to the different number of images each dataset contains, the quality of images in each dataset is also different. Images in the 291 dataset are usually small (on average, $150 \times 150$), images in ImageNet are much larger, while images in DIV2K are of very high quality. Because of the restricted resolution of the images in the 291 dataset, models on this set have difficulties in obtaining large patches with large receptive fields. Therefore, models based on the 291 dataset usually take the bicubic of LR as input, which is quite time-consuming. Table~\ref{chart} compares different models on the mentioned criteria.

\begin{table*}[!t]
\centering
\caption{Comparisons among some representative deep models.}
\begin{tabular}{ccccc}
\hline
\thead{\textbf{Models}} & \thead{\textbf{PSNR/SSIM($\times$4)}} & \thead{\textbf{Train data}} & \thead{\textbf{Parameters}} & \thead{\textbf{Mult\&Adds}}  \\
\hline
\thead{SRCNN\_EX~\cite{dong2016image}} & \thead{30.49/0.8628} & \thead{ImageNet subset} & \thead{57K} & \thead{52.5G}  \\
\thead{ESPCN~\cite{shi2016real}} & \thead{30.90/-} & \thead{ImageNet subset} & \thead{20K} & \thead{1.43G}  \\
\thead{VDSR~\cite{kim2016accurate}} & \thead{31.35/0.8838} & \thead{G200+Yang91} & \thead{665K} & \thead{612.6G}  \\
\thead{DRCN~\cite{kim2016deeply}} & \thead{31.53/0.8838} & \thead{Yang91} & \thead{1.77M(recursive)} & \thead{17974.3G}  \\
\thead{DRRN~\cite{tai2017image}} & \thead{31.68/0.8888} & \thead{G200+Yang91} & \thead{297K(recursive)} & \thead{6796.9G}  \\
\thead{LapSRN~\cite{lai2017deep}} & \thead{31.54/0.8855} & \thead{G200+Yang91} & \thead{812K} & \thead{29.9G}  \\
\thead{SRResNet~\cite{ledig2017photo}} & \thead{32.05/0.9019} & \thead{ImageNet subset} & \thead{1.5M} & \thead{127.8G}  \\
\thead{MemNet~\cite{tai2017memnet}} & \thead{31.74/0.8893} & \thead{G200+Yang91} & \thead{677K(recursive)} & \thead{2265.0G}  \\
\thead{RDN~\cite{zhang2018residual}} & \thead{32.61/0.9003} & \thead{DIV2K} & \thead{22.6M} & \thead{1300.7G}  \\
\thead{EDSR~\cite{lim2017enhanced}} & \thead{32.62/0.8984} & \thead{DIV2K} & \thead{43M} & \thead{2890.0G}  \\
\thead{MDSR~\cite{lim2017enhanced}} & \thead{32.60/0.8982} & \thead{DIV2K} & \thead{8M} & \thead{407.5G}  \\
\thead{DBPN~\cite{haris2018deep}} & \thead{32.47/0.898}  & \thead{DIV2K+Flickr+ \\ ImageNet subset} & \thead{10M} & \thead{5715.4G} \\
\hline
\end{tabular}
\label{chart}
\end{table*}

From Table~\ref{chart}, we can see that generally as the depth and the number of parameters grow, the performance improves. However, the growth rate of performance levels off. Recently, some works on designing light models~\cite{yang2017single,ahn2018fast,hui2018fast} and learning sparse structural NN~\cite{fan2018compressed} were proposed to achieve relatively good performance with less storage and computation, which are very meaningful in practice. 

\begin{figure*}[!t]
    \centering
     \subfloat[HR]{
    	\includegraphics[scale=0.25]{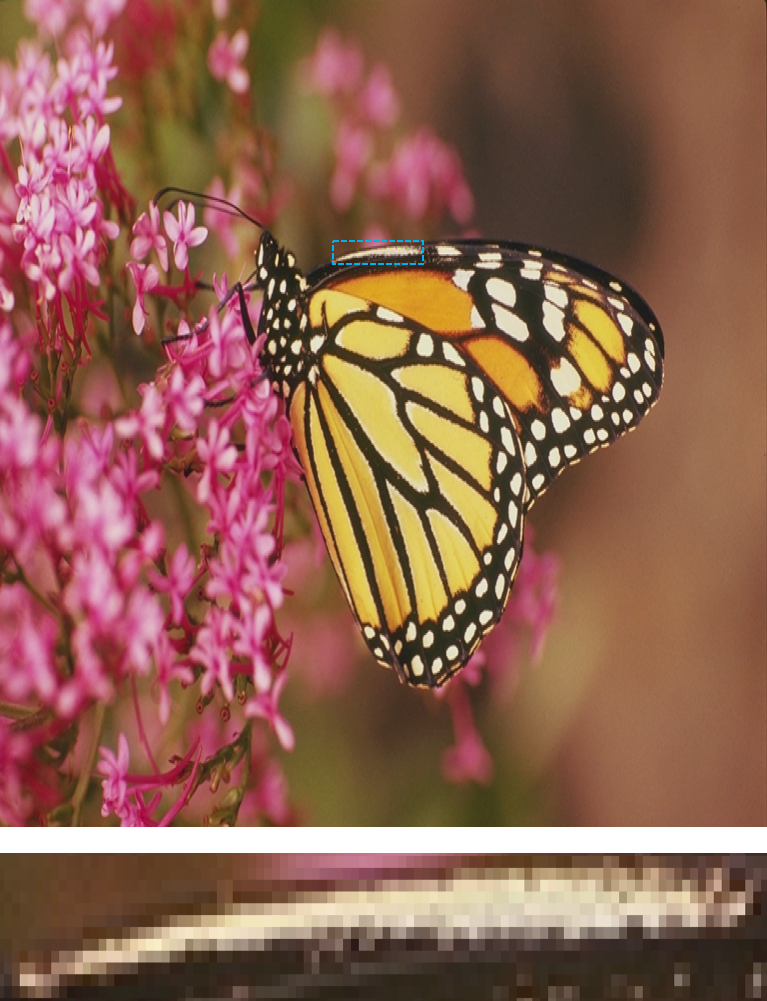}}\hfil
    \subfloat[EDSR(27.80dB/0.9012)]{
        \includegraphics[scale=0.25]{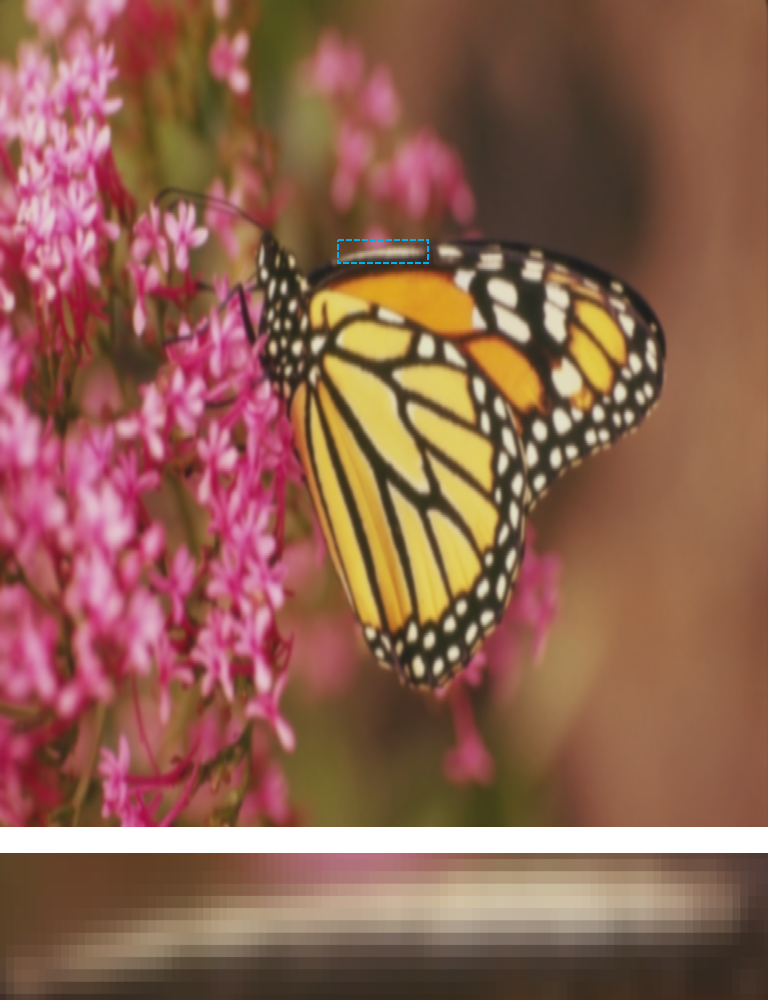}}\hfil
    \subfloat[IRCNN(34.63dB/0.9548)]{
        \includegraphics[scale=0.25]{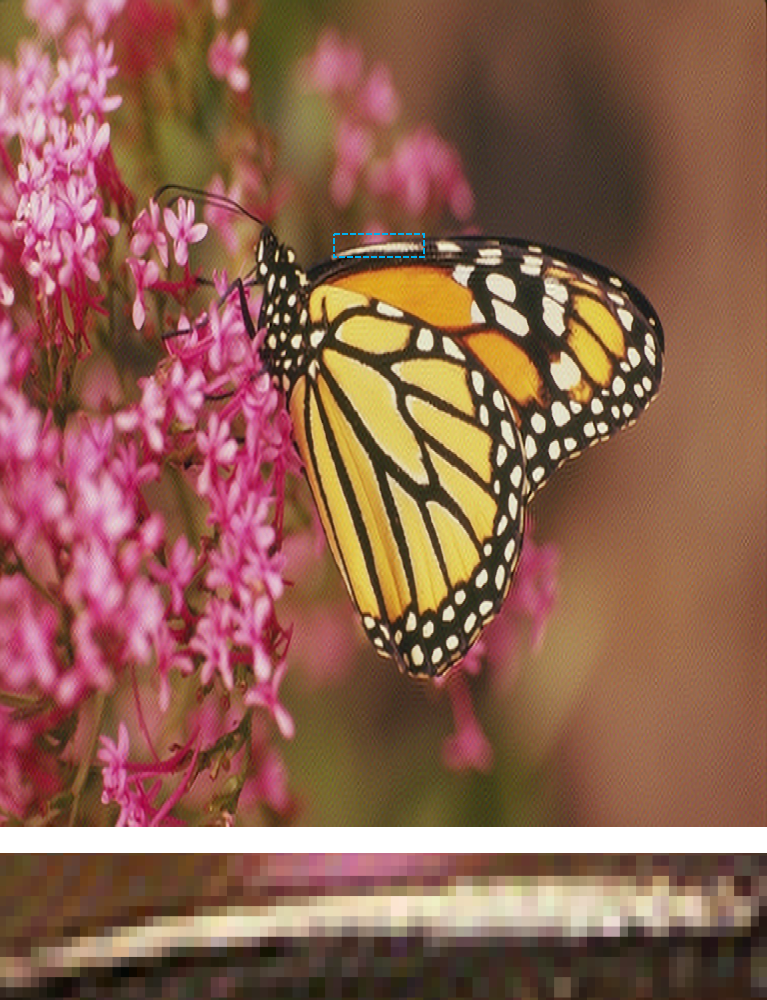}} \hfil
    \subfloat[ZSSR(30.45dB/0.9384)]{
        \includegraphics[scale=0.25]{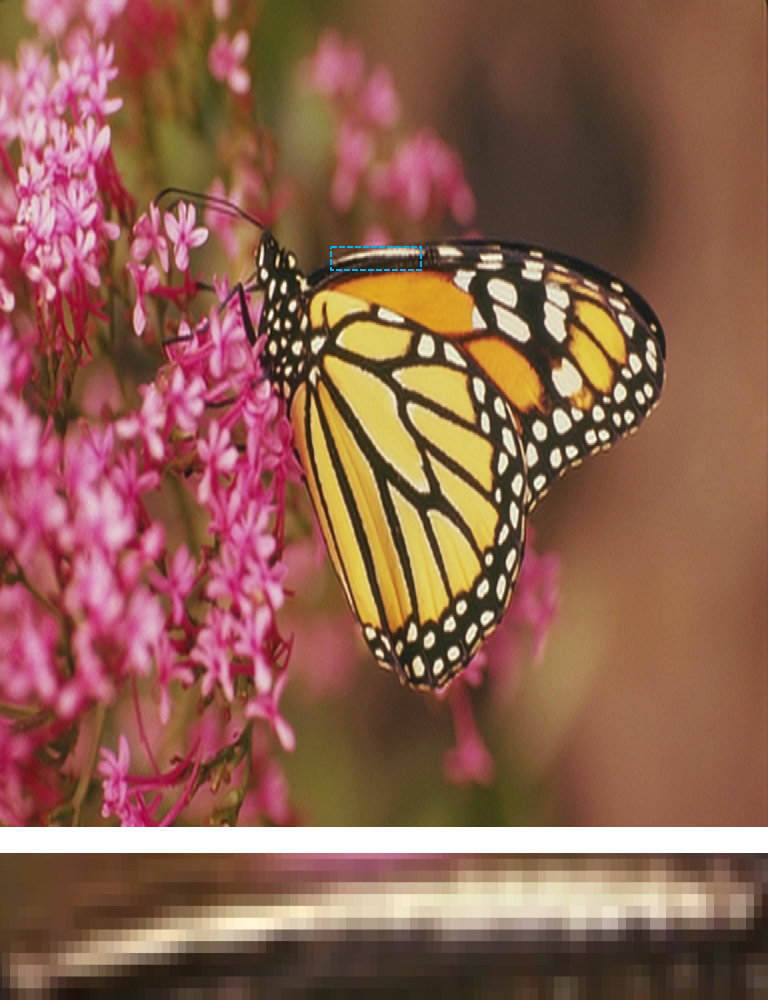}}\hfil
    \subfloat[SRDM(37.71dB/0.9723)]{
        \includegraphics[scale=0.25]{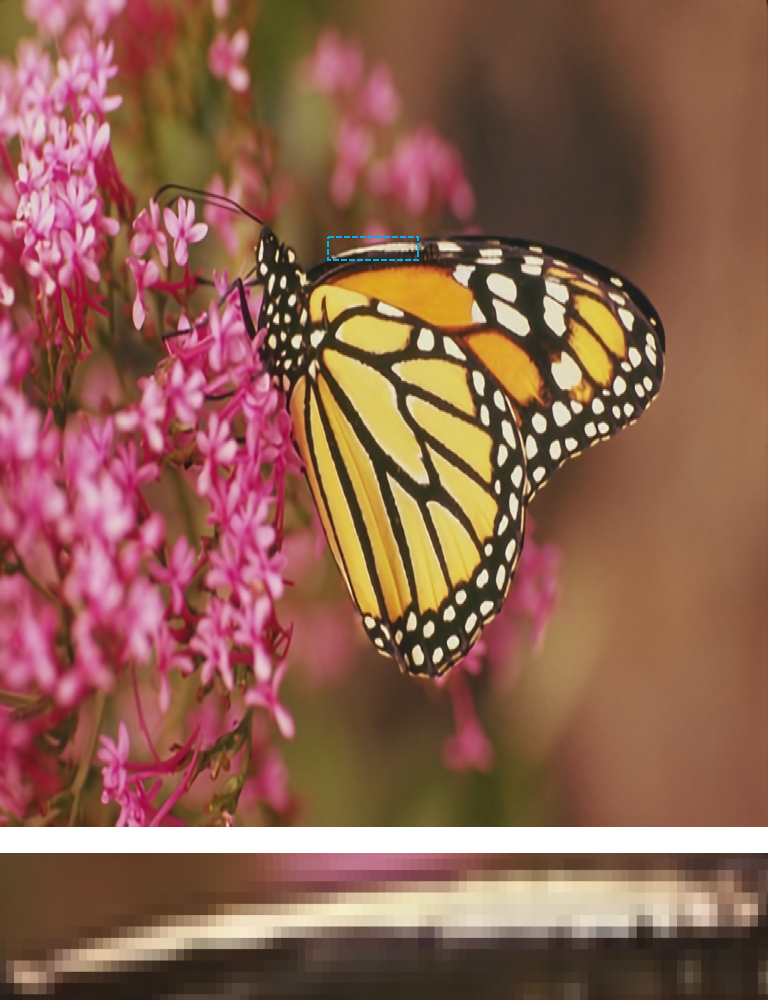}}
    \caption{Comparisons of 'monarch' in Set14 for scale 2 with Gaussian kernel degradation. We can see that, given the degradation mismatch with that of training, the performance of EDSR decreases drastically.}
    \label{fig:GaussianBlur}
\end{figure*}

For the second part, we mainly show that the performance of the models for some specific degradation dropped drastically when the true degradation mismatches the one assumed for training. For example, we use four models, including EDSR trained with bicubic degradation~\cite{lim2017enhanced}, IRCNN~\cite{zhang2017learning}, SRMD~\cite{zhang2018learning} and ZSSR~\cite{shocher2017zero}, to address LRs generated by Gaussian kernel degradation (kernel size of $7 \times 7$ with bandwidth 1.6), as shown in Fig.~\ref{fig:GaussianBlur}, and the performance of EDSR dropped drastically with obvious blur, while other models for nonspecific degradation perform quite well. Therefore, to address some longstanding problems in SISR, such as unknown degradation, the direct usage of general deep learning techniques may not be sufficient. More effective solutions can be achieved by combining the power of DL and the specific properties of the SISR scene.

\section{Optimization Objectives for DL-based SISR}\label{s:obj}

\subsection{Benchmark of Optimization Objectives for DL-based SISR}
We select the MSE loss used in SRCNN as the benchmark. It is known that using MSE favors a high PSNR, and PSNR is a widely used metric for quantitatively evaluating image restoration quality. Optimizing MSE can be viewed as a regression problem, leading to a point estimation of $\theta$ as
\begin{equation}
\begin{split}
\min \limits_{\theta} \sum_{i}||F(x_{i}; \theta)-y_i||^{2},
\end{split}
\label{mse}
\end{equation}
where $(x_{i}, y_{i})$ are the $i$th training examples and $F(x;\theta)$ is a CNN parameterized by $\theta$. Here, (\ref{mse}) can be interpreted in a probabilistic way by assuming Gaussian white noise ($\mathcal N(\epsilon; 0, \sigma^{2}I)$) independent of the image in the regression model, and then, the conditional probability of $y$ given $x$ becomes a Gaussian distribution with mean $F(x;  \theta)$ and the diagonal covariance matrix $\sigma^2 I$, where $I$ is the identity matrix:  
\begin{equation}
\begin{split}
p(y|x)=\mathcal  N(y;F(x;\theta), \sigma^{2}I).
\end{split}
\label{conditionprob}
\end{equation}
Then, using maximum likelihood estimation (MLE) on the training examples with (\ref{conditionprob}) will lead to (\ref{mse}).

The Kullback-Leibler divergence (KLD) between the conditional empirical distribution $P_{data}$ and the conditional model distribution $P_{model}$ is defined as 
\begin{equation}
\begin{split}
D_{KL}(P_{data}||P_{model})=E_{z\sim P_{data}}[\log \frac{P_{data}(z)} {P_{model}(z)}].
\end{split}
\label{KL_former}
\end{equation}
We call (\ref{KL_former}) the forward KLD, where $z=y|x$ denotes the HR (SR) conditioned on its LR counterpart, $P_{data}$ and $P_{model}$ are the conditional distributions of $HR|LR$ and $SR|LR$, respectively, where $E_{x\sim P_{data}}[\log {P_{data}(z)}]$ is an intrinsic term determined by the training data regardless of the parameter $\theta$ of the model (or the model distribution $P_{model}(x; \theta)$).  Hence, when we use the training samples to estimate parameter $\theta$, minimizing this KLD is equivalent to MLE. 

Here, we have demonstrated that MSE is a special case of MLE, and MLE is a special case of KLD. However, we may conjecture whether the assumptions underlying these specializations are violated. This consideration has led to some emerging objective functions from four perspectives:

1) Translating MLE into MSE can be achieved by assuming Gaussian white noise. Although the Gaussian model is the most widely used model for its simplicity and technical support, what if this independent Gaussian noise assumption is violated in a complicated scene such as SISR?

2) To use MLE, we need to assume the parametric form of the data distribution. What if the parametric form is misspecified?

3) Apart from KLD in (\ref{KL_former}), are there any other distances between probability measures that we can use as the optimization objectives for SISR? 

4) Under specific circumstances, how can we choose the suitable objective functions according to their properties?

Based on some solutions to these four questions, recent work on optimization objectives for DL-based SISR will be discussed in Sections~\ref{ss:nonGaussian}, \ref{ss:nonparam}, \ref{ss:otherdistance} and \ref{ss:different}, respectively.

\subsection{Objective Functions Based on non-Gaussian Additive Noises}\label{ss:nonGaussian}

The poor perceptual quality of the SISR images obtained by optimizing MSE directly demonstrates a fact: using Gaussian additive noise in the HR space is not good enough. To address this problem, solutions are proposed from two aspects: use other distributions for this additive noise, or transfer the HR space to some space where the Gaussian noise is reasonable. 

\subsubsection{Denote Additive Noise with Other Probability Distributions}
In~\cite{Zhao2015Loss}, Zhao \emph{et~al.} investigated the difference between mean absolute error (MAE) and MSE used to it optimize NN in image processing. Similar to (\ref{mse}), MAE can be written as 
\begin{equation}
\begin{split}
\min \limits_{\theta} \sum_{i}||F(x_{i}; \theta)-y_i||_{1}.
\end{split}
\label{mae}
\end{equation}
From the perspective of probability, (\ref{mae}) can be interpreted as introducing Laplacian white noise, and similar to (\ref{conditionprob}), the conditional probability becomes
\begin{equation}
\begin{split}
p(y|x)=Laplace(y;F(x;\theta), bI).
\end{split}
\label{laplace}
\end{equation} 
Compared with MSE in regression, MAE is believed to be more robust against outliers. As reported in~\cite{Zhao2015Loss}, when MAE is used to optimize an NN, the NN tends to converge faster and produce better results. The authors argued that the reason might be because MAE could guide NN to reach a better local minimum. Other similar loss functions in robust statistics can be viewed as modeling additive noises with other probability distributions. 

Although these specific distributions often cannot represent unknown additive noise very precisely, their corresponding robust statistical loss functions are used in many DL-based SISR works for their conciseness and advantages over MSE.

\subsubsection{Using MSE in a Transformed Space}
Alternatively, we can search for a mapping $\Psi$ to transform the HR space to some space where Gaussian white noise can be used reasonably.
From this perspective, Bruna \emph{et~al.}~\cite{bruna2015super} proposed so-called perceptual loss to leverage deep architectures. In~\cite{bruna2015super}, the conditional probability of the residual  $r$ between HR and LR given the LR $x$ is stimulated by the Gibbs energy model:
\begin{equation}
\begin{split}
p(r|x) = \exp(-{||\Phi(x)-\Psi(r)||}^{2}-\log Z),
\end{split}
\label{Gibbs}
\end{equation}
where $\Phi$ and $\Psi$ are two mappings between the original spaces and the transformed ones, and $Z$ is the partition function. The features produced by sophisticated supervised deep architectures have been shown to be perceptually stable and discriminative, denoted by $\Psi(r)$\footnote{Either the scattering network or VGG can be denoted by $\Psi$. When $\Psi$ is VGG, there is no residual learning and fine-tuning.}. Then, $\Psi$ represents the corresponding deep architectures. In contrast, $\Phi$ is the mapping between the LR space and the manifold represented by $\Psi(r)$, trained by minimizing the Euclidean distance as
\begin{equation}
\begin{split}
\min \limits_{\Phi}{||\Phi(x)-\Psi(r)||}^2.
\end{split}
\label{phi}
\end{equation}
After $\Phi$ is obtained, the final result $r$ can be inferred with SGD by solving

\begin{equation}
\begin{split}
\min \limits_{r}{||\Phi(x)-\Psi(r)||}^2.
\end{split}
\label{infer}
\end{equation}

For further improvement, \cite{bruna2015super} also proposed a fine-tuning algorithm in which $\Phi$ and $\Psi$ can be fine-tuned to the data. Similar to the alternating updating in GAN, $\Phi$ and $\Psi$ are fine-tuned with SGD based on the current $r$. However, this fine-tuning will involve calculating the gradient of the partition function $Z$, which is a well-known difficult decomposition into the positive phase and the negative phase of learning. Hence to avoid sampling within inner loops, a biased estimator of this gradient is chosen for simplicity.

The inference algorithm in~\cite{bruna2015super} is extremely time-consuming. To improve efficiency, Johnson \emph{et~al.} utilized this perceptual loss in an end-to-end training manner~\cite{johnson2016perceptual}. In~\cite{johnson2016perceptual}, the SISR network is directly optimized with SGD by minimizing the MSE in the feature manifold produced by VGG-16 as follows: 
\begin{equation}
\begin{split}
\min \limits_{\theta}{||\Psi(F(x;\theta))-\Psi(y)||}^2,
\end{split}
\label{ECCV16}
\end{equation}
where $\Psi$ is the mapping represented by VGG-16, $F(x;\theta)$ denotes the SISR network, and $y$ is the ground truth. Compared with~\cite{bruna2015super},~\cite{johnson2016perceptual} replaces the nonlinear mapping $\Phi$ and the expensive inference with an end-to-end trained CNN, and their results show that this change does not affect the restoration quality but does accelerate the whole process. 

Perceptual loss mitigates blurring and leads to more visually-pleasing results compared with directly optimizing MSE in the HR space. However, there remains no theoretical analysis on why this approach  works. In~\cite{bruna2015super}, the author generally concluded that successful supervised networks used for high-level tasks could produce very compact and stable features. In these feature spaces, small pixel-level variation and much other trivial information can be omitted, making these feature maps mainly focus on pixels of human interest. At the same time, with the deep architectures, the most specific and discriminative information of the input is shown to be retained in feature spaces because of the great performance of the models applied in various high-level tasks. From this perspective, using MSE in these feature spaces will focus more on the parts that are attractive to human observers with little loss of original contents, so perceptually pleasing results can be obtained. 

\subsection{Optimizing Forward KLD with Nonparametric Estimation}\label{ss:nonparam}

Parametric estimation methods such as MLE need to specify in advance the parametric form the distribution of data, which suffers from model misspecification. Different from parametric estimation, nonparametric estimation methods such as kernel distribution estimation (KDE) fit the data without distributional assumptions, which are robust when the real distributional form is unknown. Based on nonparametric estimation, recently, the contextual loss~\cite{mechrez2018learning,mechrez2018contextual} was proposed by Mechrez \emph{et~al.} to maintain natural image statistics. In the contextual loss, a Gaussian kernel function is applied:
\begin{equation}
\begin{split}
K(x,y)= \exp(-dist(x,y)/h - \log Z),
\end{split}
\label{kernelfunction}
\end{equation}
where $dist(x,y)$ can be any symmetric distance between $x$ and $y$, $h$ is the bandwidth, and the partition function $Z=\int \exp(-dist(x,y)/h)dy$. Then, $P_{data}$ and $P_{model}$ are
\begin{equation}
\begin{split}
P_{data}(z)=\sum_{z_{i}\sim P_{data}}K(z,z_{i}), \\ P_{model}(z)=\sum_{w_{j}\sim P_{model}}K(z,w_{j}),
\end{split}
\label{kenerlestimate}
\end{equation} 
and (\ref{KL_former}) can be rewritten as
\begin{equation}
\begin{split}
&D_{KL}(P_{data}||P_{model})=\\
&\frac{1}{N}\sum_{k}[\log \sum_{z_{i}\sim P_{data}}K(z_{k},z_{i})-\log \sum_{w_{j}\sim P_{model}}K(z_{k},w_{j})].
\end{split}
\label{KL_KDE}
\end{equation}

The first log term in (\ref{KL_KDE}) is a constant with respect to the model parameters. Let us denote the kernel $K(z_{k},w_{j})$ in the second log term by $A_{kj}$.  Then, the optimization objective in (\ref{KL_KDE}) can be rewritten as
\begin{equation}
-\frac{1}{N}\sum_{k}\log \sum_{j} A_{kj}.
\label{cross_entropy}
\end{equation}
With the Jensen inequality, we can obtain a lower bound of (\ref{cross_entropy}):
\begin{equation}
-\frac{1}{N}\sum_{k}\log \sum_{j} A_{kj} \ge -\log \frac{1}{N}\sum_{k}\sum_{j}A_{kj} \ge 0.
\label{CE_lower}
\end{equation}
The first equality holds if and only if $\forall k, k'$, $\sum_{j}A_{kj}=\sum_{j}A_{k'j}$. Both equalities hold if and only if $\forall k,~\sum_{j}A_{kj}=0$. When (\ref{cross_entropy}) reaches 0, the given lower bound also reaches 0. Therefore, we can take this lower bound as the optimization objective alternatively.

We can further simplify the lower bound in (\ref{CE_lower}). The lower bound can be rewritten as
\begin{equation}
-\log \frac{1}{N}\sum_{j}||A_{j}||_{1},
\label{CE_lower_trans}
\end{equation}
where $A_{j}=(A_{1j},\cdots,A_{kj})^{T}$, and ${\| \cdot \|}_{1}$ is the $\ell_{1}$ norm. When the bandwidth $h \to 0$, the affinity $A_{kj}$ will degrade into the indicator function, which means if $x_{k}=y_{j}$, $A_{kj}\approx 1$; otherwise, $A_{kj}\approx 0$. In this case, the $\ell_{1}$ norm can be approximated well by the $\ell_{\infty}$ norm, which returns the maximum element of the vector. Thus, (\ref{CE_lower_trans}) can degenerate into the contextual loss in~\cite{mechrez2018learning,mechrez2018contextual}:
\begin{equation}
-\log \frac{1}{N}\sum_{j}\max_{k}A_{kj}.
\label{contextual_loss}
\end{equation}

Recently, implicit likelihood estimation (IMLE)~\cite{li2018implicit} was proposed and its conditional version was applied to SISR~\cite{li2018super}. Here, we will briefly show that minimizing IMLE equals minimizing an upper bound of the forward KLD  with KDE. Let us use a Gaussian kernel as
\begin{equation}
    \begin{split}
        K(x,y)=\frac{1}{\sqrt{2\pi}h}\exp\left(-\frac{\|x-y\|_{2}^{2}}{2h^2}\right).
    \end{split}
    \label{RBF}
\end{equation}

As with (\ref{cross_entropy}), the optimization objective can be rewritten as 
\begin{equation}
    \begin{split}
        -\frac{1}{N}\sum_{k} \log \sum_{j} e^{-\frac{\|z_k-w_j\|_2^2}{2h^2}}.
    \end{split}
    \label{new_cross_entropy}
\end{equation}
With $\{w_j\}_{j=1}^m$ and $\{z_k\}_{k=1}^N$, we can obtain a simple upper bound of (\ref{new_cross_entropy}) as
\begin{equation}
    \begin{split}
        -\frac{1}{N}\sum_{k} \log \left(m \min_{j}
        e^{-\frac{\|z_k-w_j\|_2^2}{2h^2}}\right)\\
        =\frac{1}{N}\sum_k(\min_j\frac{||z_k-w_j||_2^2}{2h^2}-\log m).
    \end{split}
    \label{ce_upperbound}
\end{equation}
Minimizing (\ref{ce_upperbound}) equals minimizing 
\begin{equation}
    \begin{split}
        \sum_k \min_j \|z_k-w_j\|_2^2,
    \end{split}
    \label{IMLE}
\end{equation}
which is the core of the optimization objective of IMLE.

As above, the recently proposed contextual loss and IMLE are illustrated via nonparametric estimation and KLD.  
Visually pleasing results were reported using the contextual loss and IMLE. 
However, as KDE is generally very time-consuming, several reasonable approximations along with acceleration algorithms were applied.

\subsection{Other Distances between Probability Measures Used in SISR}\label{ss:otherdistance}

\begin{figure}[t] 
    \centering
    \subfloat[forward KLD]{
       \includegraphics[scale=0.35]{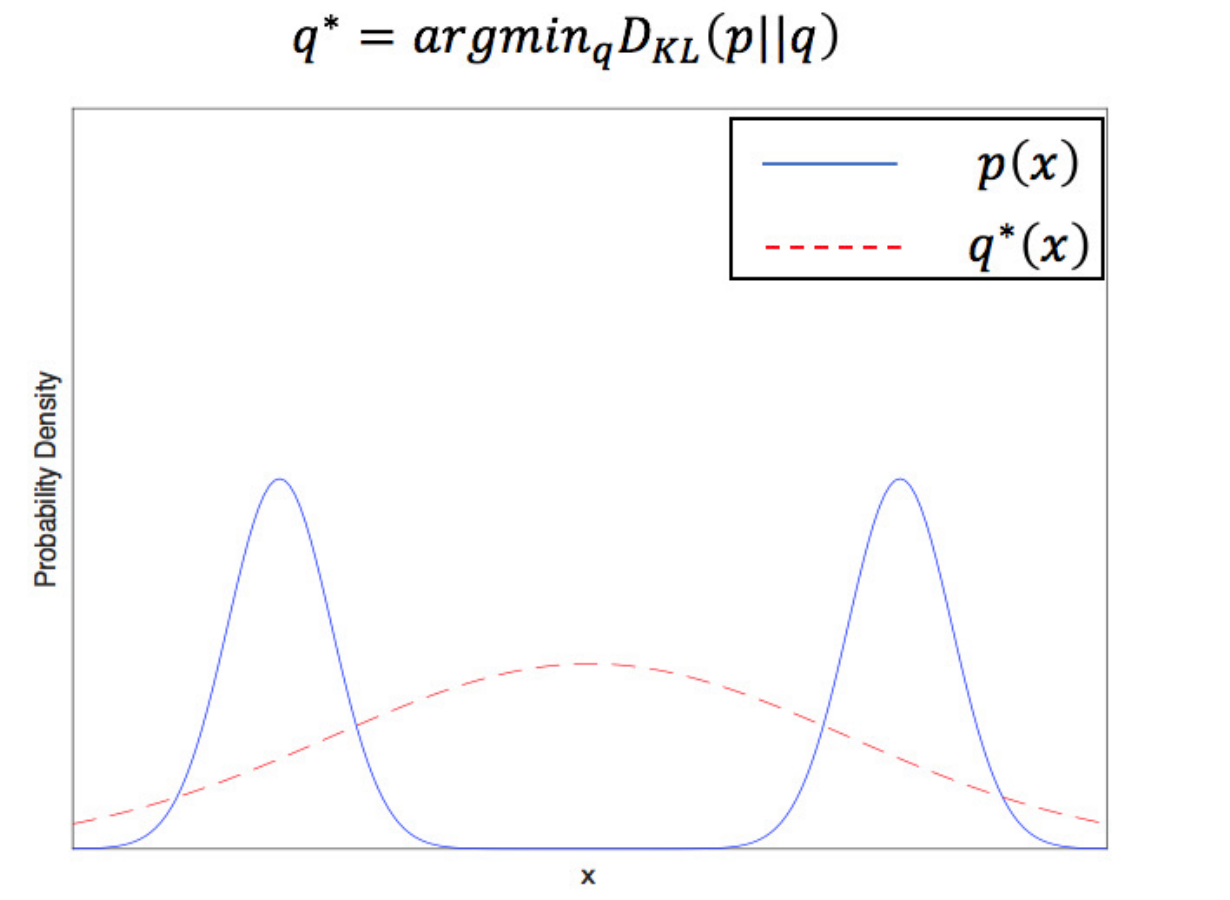}}
    \label{forward}\hfill
    \subfloat[backward KLD]{
        \includegraphics[scale=0.35]{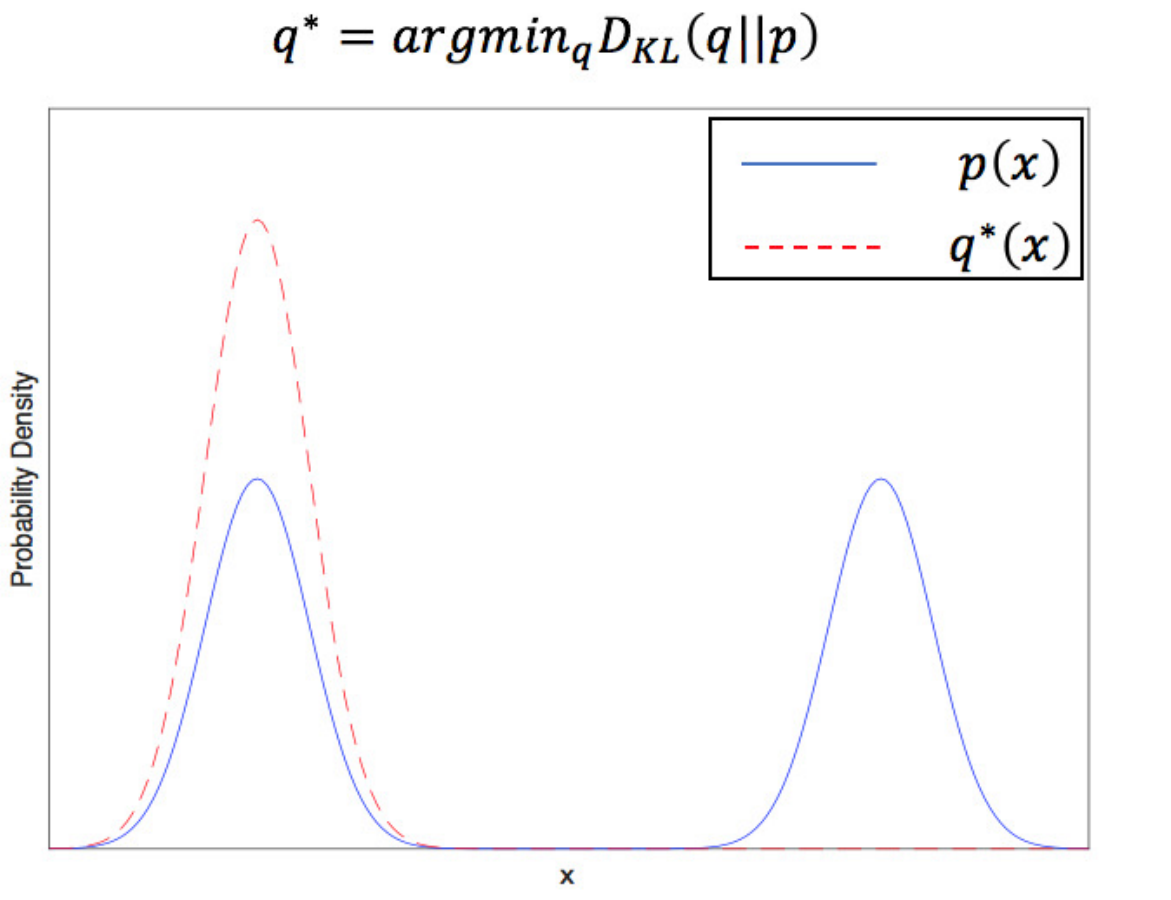}}
    \label{backward}\hfill
    \caption{A toy example to illustrate the difference between the forward KLD and the backward KLD.}
    \label{KLD} 
\end{figure}

As KLD is an asymmetric (pseudo) distance for measuring similarity between two distributions, in this subsection, we begin with the inverse form of forward KLD, namely, backward KLD. The backward KLD is defined as 
\begin{equation}
\begin{split}
D_{KL}(P_{model}||P_{data})=E_{z\sim P_{model}}[\log \frac{P_{model}(z)}{P_{data}(z)}].
\end{split}
\label{KL_latter}
\end{equation}
When $P_{model}=P_{data}$, both KLDs reach the minimum of 0. However, when the solution is inadequate, these two KLDs will lead to quite different results. Here, we use a toy example to illustrate a simple case of inadequate solutions, as shown in Fig.~\ref{KLD}. 

The unknown wanted distribution is a Gaussian mixture model (GMM) with two modes, denoted as $P(x)$, and we model it by a single Gaussian distribution. We can easily see that optimizing the forward KLD results in a solution locating at the middle areas of two modes, while optimizing the backward KLD makes the result close to the most prominent mode.

From Fig.~\ref{KLD} we can see that, under inadequate solutions, optimizing the forward KLD will lead to the well-known regression-to-the-mean problem, while optimizing the backward KLD only concentrates on the main modality. The former is one of the reasons for blurring, and some researchers \cite{huszar2015not} argued that the latter improves the visual quality but makes the results collapse to some patterns.

Different distances may lead to different results under an inadequate solution. Readers can refer to~\cite{theis2015note} for further understanding. In most low-level computer vision tasks, $P_{data}$ is an empirical distribution and $P_{model}$ is an intractable distribution. For this reason, the backward KLD is unpractical for optimizing deep architectures. To relieve optimizing difficulties, we replace the asymmetric KLD with the symmetric Jensen-Shannon divergence (JSD) as follows: 
\begin{equation}
\begin{split}
JS(P_{data}||P_{model})= & \frac{1}{2} KL[P_{data}||\frac{P_{data}+P_{model}}{2}]+ \\ &\frac{1}{2} KL[P_{model}||\frac{P_{data}+P_{model}}{2}].
\end{split}
\label{JSD}
\end{equation}

\begin{figure*}
	\centering
    \subfloat[\footnotesize HR]{
    	\includegraphics[scale=0.35]{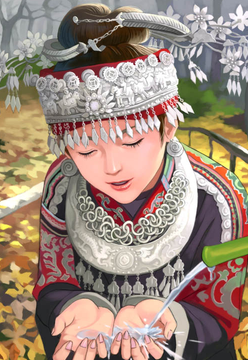}}
    \label{comic_HR}\hfill
    \subfloat[\footnotesize bicubic(21.59dB/0.6423)]{
    	\includegraphics[scale=0.35]{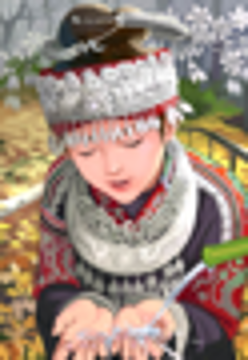}}
    \label{comic_bicubic}\hfill
    \subfloat[\footnotesize SRResNet(23.53dB/0.7832)]{
    	\includegraphics[scale=0.35]{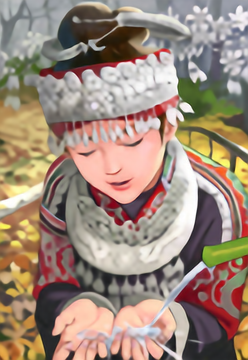}}
    \label{comic_MSE}\hfill
    \subfloat[\footnotesize SRGAN(21.15dB/0.6868)]{
    	\includegraphics[scale=0.35]{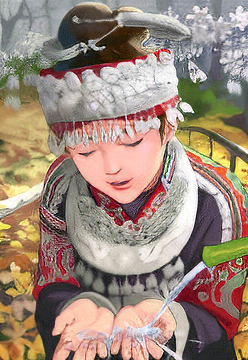}}
    \label{comic_GAN}\hfill
    \subfloat[\footnotesize SRCX(20.88dB/0.6002)]{
        \includegraphics[scale=0.35]{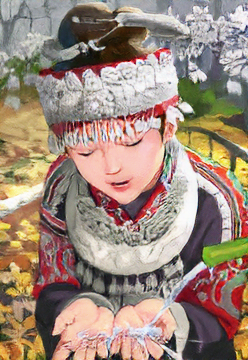}}
    \label{comic_CX}
    \caption{Visual comparisons between the MSE, MSE + GAN and MAE  +GAN + Contextual Loss (The authors of~\cite{ledig2017photo} and~\cite{mechrez2018contextual} released their results.) We can see that the perceptual loss leads to a lower PSNR/SSIM but a better visual quality.}
    \label{GAN_MSE}
\end{figure*}

Optimizing (\ref{JSD}) explicitly is also very difficult. Generative adversarial nets (GANs) proposed by Goodfellow \emph{et~al.} use the objective function below to implicitly address this problem in a game theory scenario, successfully avoiding the troubling approximate inference and approximation of the partition function gradient:
\begin{equation}
\begin{split}
\min \limits_{G} \max \limits_{D}[E_{z\sim P_{data}}\log D(z)+E_{z\sim P_{model}}\log (1-D(z))],
\end{split}
\label{GAN}
\end{equation}
where $G$ is the main part called the generator supervised by an auxiliary part $D$ called the discriminator. The two parts update alternatively, and when the discriminator cannot give useful information to the generator anymore, in other words, the outputs of the generator totally confuse the discriminator, the optimization procedure is completed. For the detailed discussion on GANs, readers can refer to~\cite{goodfellow2014generative}. Recent works have shown that sophisticated architectures and suitable hyperparameters can help GANs perform excellently. The representative works on GAN-based SISR are~\cite{ledig2017photo} and~\cite{sajjadi2017enhancenet}. In~\cite{ledig2017photo}, the generator of the GAN is the SRResNet mentioned previously, and the discriminator refers to the design criterion of DCGAN~\cite{radford2015unsupervised}. In the context of GANs, a recent work~\cite{sajjadi2017enhancenet} follows a similar path except with a different architecture. Very recently, by leveraging the extension of the basic GAN framework~\cite{zhu2017unpaired}, \cite{yuan2018unsupervised} was proposed as an unsupervised SR algorithm. Fig.~\ref{GAN_MSE} shows the results of the GAN and MSE with the same architecture; despite the lower PSNR due to artifacts, the visual quality improves by using the GAN for SISR.

Generally, GANs offer an implicit optimization strategy in an adversarial training way by using deep neural networks. Based on this, more rational but complicated measures such as Wasserstein distances~\cite{arjovsky2017wasserstein}, $f$-divergence~\cite{nowozin2016f}\footnote{Forward KLD, backward KLD and JSD can all be regarded as the special cases of $f$-divergence.} and maximum mean discrepancy (MMD)~\cite{sutherland2016generative} are taken as alternatives to JSD for training GANs.    

\subsection{Characters of Different objective functions}\label{ss:different}

Now, we can see that those losses mentioned in Section \ref{ss:nonGaussian} explicitly model the relation between LR and its HR counterpart. Here, we follow the methodology of~\cite{blau2018perception} and call the losses that were based on measuring the dissimilarity between training pairs the distortion-aimed losses. When the training data are not sufficient, distortion losses usually ignore the particularity of data and appear ineffective to measure the similarity between the source and target distributions. 

\begin{figure}
    \centering
    \subfloat[Perception-distortion trade-off]{
        \includegraphics[scale=0.25]{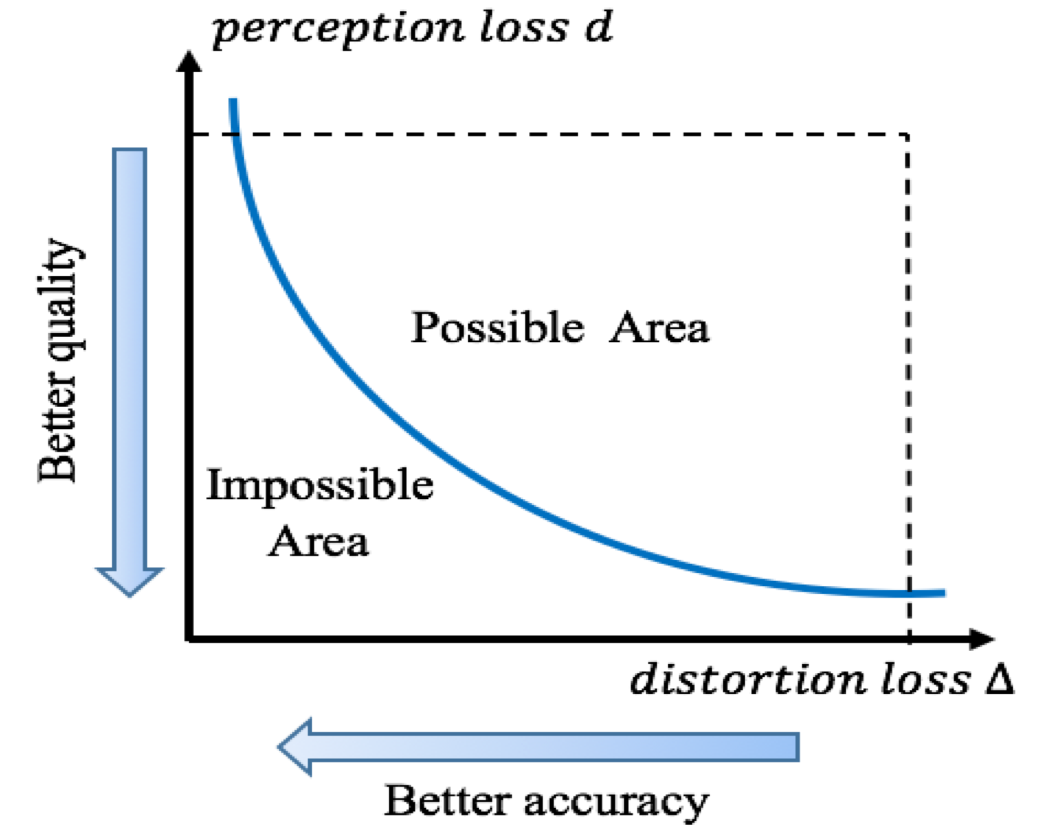}}
    \subfloat[Perception-distortion evaluation]{
        \includegraphics[scale=0.25]{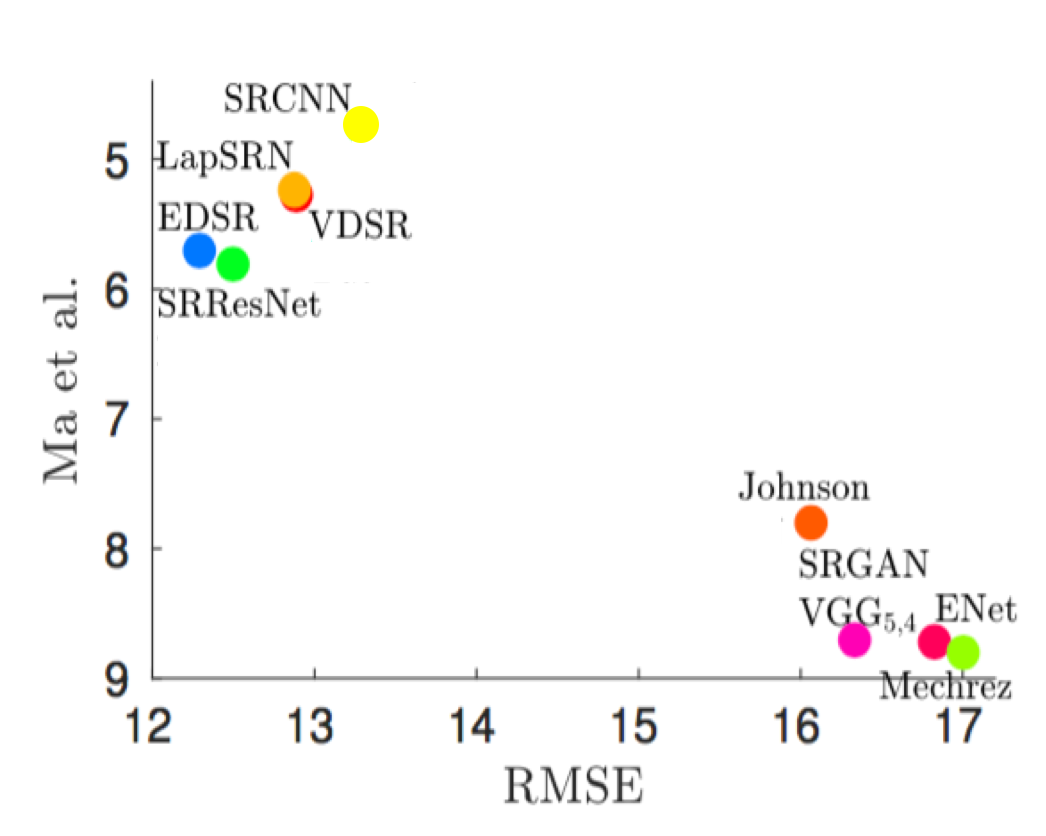}}
    \caption{(a) The perception-distortion space is divided by the perception-distortion curve, where an area cannot be attained. (b) Use of the nonreference metric proposed by~\cite{meinhardt2017learning} and RMSE to perform quantitative comparisons from the perception and distortion views; the included methods are~\cite{dong2014learning, lai2017deep,lim2017enhanced,kim2016accurate,ledig2017photo,sajjadi2017enhancenet,mechrez2018contextual,johnson2016perceptual}.}
    \label{fig:tradeoff}
\end{figure}

The losses mentioned in Sections~\ref{ss:nonparam} and \ref{ss:otherdistance} are rooted from measuring the similarity between distributions, which is thought to measure the perceptual quality. Here, we call them perception-aimed losses. Recently, Blau \emph{et~al.}~\cite{blau2018perception} discussed the inherent trade-off between the two kinds of losses. Their discussion can be simplified into an optimization problem:
\begin{equation}
    \begin{split}
        P(D) = \min \limits_{P_{\hat{Y}|X}} d(P_{Y}, P_{\hat{Y}})~s.t.~E[\Delta(Y, \hat{Y})] \le D.
    \end{split}
    \label{tradeoff}
\end{equation}
$\Delta(\cdot, \cdot)$ is distortion-aimed loss, and $d(\cdot, \cdot)$ is the (pseudo) distance between distributions. Furthermore, the author also proved that if $d(\cdot, \cdot)$ is convex in its second argument, then the $P(D)$ is monotonically nonincreasing and convex. From this property, we can draw the curve of $P(D)$ and easily see this trade-off, as shown in Fig.~\ref{fig:tradeoff}(a), such that improving one must be at the expense of the other. However, as shown in Section~\ref{ss:nonGaussian}, using MSE in the VGG feature space achieves a better quality, and choosing suitable $\Delta$ and $d$ may ease this trade-off. 

For the perception-aimed losses mentioned in Sections~\ref{ss:nonparam} and~\ref{ss:otherdistance}, up to now, there has been no rigorous analysis on their differences. Here, we apply the nonreference quality assessment proposed by Ma \emph{et~al.}~\cite{meinhardt2017learning} with RMSE to conduct quantitative comparisons, and the representative qualitative comparisons are depicted in Fig.~\ref{fig:tradeoff}(b). To summarize, we should be aware that there is no one-fits-all objective function, and we should choose one that is suitable to the context of an application.

\section{Trends and Challenges}\label{s:trend}

Along with the promising performance that DL algorithms have achieved in SISR, there remain several important challenges and inherent trends as follows.

\subsubsection{Lighter Deep Architectures for Efficient SISR}
Although the high accuracy of advanced deep models has been achieved for SISR, it is still difficult to deploy these models to real-world scenarios, which is mainly due to massive parameters and computation. To address this issue, we need to design light deep models or slim the existing deep models for SISR with fewer parameters and computation at the expense of little or no performance degradation. Hence, in the future, researchers are expected to focus more on reducing the size of NNs for speeding up the SISR process.

\subsubsection{More Effective DL Algorithms for Large-scale SISR and SISR with Unknown Corruption}
Generally, DL algorithms proposed in recent years have improved the performance of traditional SISR tasks by a large margin. However, the large scale of SISR and the SISR with unknown corruption, the two major challenges in the SR community, are still lacking very effective remedies. DL algorithms are thought to be skilled at addressing many inferences or unsupervised problems, which is of key importance to address these two challenges. Therefore, by leveraging the great power of DL, more effective solutions to these two demanding problems are expected.

\subsubsection{Theoretical Understanding of Deep Models for SISR}
The success of deep learning is said to be attributed to learning powerful representations. However, to date, we still cannot understand these representations very well, and the deep architectures are treated as a black box. For DL-based SISR, the deep architectures are often viewed as a universal approximation, and the learned representations are often omitted for simplicity. This behavior is not beneficial for further exploration. Therefore, we should not only focus on whether a deep model works but also concentrate on why and how it works. That is, more theoretical explorations are needed.

\subsubsection{More Rational Assessment Criteria for SISR in Different Applications}
In many applications, we need to design the desired objective function for a specific application. However, in most cases, we cannot give an explicit and precise definition to assess the requirement for the application, which leads to the vagueness of the optimization objectives. Many works, although for different purposes, simply employ MSE as the assessment criterion, which has been shown as a poor criterion in many cases. In the future, we think that it is of great necessity to make clear definitions for assessments in various applications. Based on these criteria, we can design better targeted optimization objectives and compare algorithms in the same context more rationally.

\section{Conclusion}\label{s:concl}
This paper presents a brief review of recent deep learning algorithms on SISR. It divides the recent works into two categories: the deep architectures for simulating the SISR process and the optimization objectives for optimizing the whole process. Despite the promising results reported so far, there are still many underlying problems. We summarize the main challenges into three aspects: the acceleration of deep models, the extensive comprehension of deep models and the criteria for designing and evaluating the objective functions. Along with these challenges, several directions may be further explored in the future.

\section*{Acknowledgment}
We are grateful to the authors of~\cite{dong2014learning, lai2017deep,lim2017enhanced,kim2016accurate,ledig2017photo,sajjadi2017enhancenet,mechrez2018contextual,johnson2016perceptual,zhang2017learning,zhang2018learning,shocher2017zero} for kindly releasing their experimental results or codes, as well as to the three anonymous reviewers for their constructive criticism, which has significantly improved our manuscript. Moreover, we thank Qiqi Bao for helpful discussions. 

\bibliographystyle{IEEEtran}
\bibliography{review}

\end{document}